\documentclass[english,twoside]{article}
\usepackage[T1]{fontenc}
\usepackage[latin9]{inputenc}
\usepackage{color}
\usepackage{babel}
\usepackage{float}
\usepackage{mathtools}
\usepackage{dsfont}
\usepackage{amsmath}
\usepackage{amsthm}
\usepackage{amssymb}
\usepackage[authoryear,round]{natbib}
\usepackage[unicode=true,pdfusetitle,
 bookmarks=true,bookmarksnumbered=false,bookmarksopen=false,
 breaklinks=false,pdfborder={0 0 1},backref=false,colorlinks=false]
 {hyperref}

\makeatletter

\floatstyle{ruled}
\newfloat{algorithm}{tbp}{loa}
\providecommand{\algorithmname}{Algorithm}
\floatname{algorithm}{\protect\algorithmname}

\theoremstyle{plain}
\newtheorem{thm}{\protect\theoremname}[section]
\theoremstyle{definition}
\newtheorem{defn}[thm]{\protect\definitionname}
\theoremstyle{plain}
\newtheorem{lem}[thm]{\protect\lemmaname}
\theoremstyle{plain}
\newtheorem{cor}[thm]{\protect\corollaryname}
\ifx\proof\undefined
\newenvironment{proof}[1][\protect\proofname]{\par
	\normalfont\topsep6\p@\@plus6\p@\relax
	\trivlist
	\itemindent\parindent
	\item[\hskip\labelsep\scshape #1]\ignorespaces
}{%
	\endtrivlist\@endpefalse
}
\providecommand{\proofname}{Proof}
\fi

\usepackage[accepted]{arxiv}

\makeatother

\providecommand{\corollaryname}{Corollary}
\providecommand{\definitionname}{Definition}
\providecommand{\lemmaname}{Lemma}
\providecommand{\theoremname}{Theorem}

\begin{document}
\global\long\def\E{\mathbb{\mathbb{E}}}%
\global\long\def\F{\mathcal{F}}%
\global\long\def\R{\mathbb{R}}%
\global\long\def\algname{\text{Generalized SignSTORM}}%
\global\long\def\domxi{\mathcal{D}}%
\global\long\def\bL{\mathbb{\mathbf{\mathbf{L}}}}%
\global\long\def\bzero{\mathbb{\mathbf{\mathbf{0}}}}%
\global\long\def\sgn{\text{sgn}}%
\global\long\def\bs{\boldsymbol{\sigma}}%
\global\long\def\pa{\partial}%
\global\long\def\na{\nabla}%
\global\long\def\N{\mathbb{N}}%

\runningtitle{Near-Optimal Non-Convex Stochastic Optimization under Generalized Smoothness}
\twocolumn[
\aistatstitle{Near-Optimal Non-Convex Stochastic Optimization \\ under Generalized Smoothness}
\aistatsauthor{ Zijian Liu \And Srikanth Jagabathula \And  Zhengyuan Zhou }
\aistatsaddress{ Stern School of Business, New York University }]
\begin{abstract}
The generalized smooth condition, $(L_{0},L_{1})$-smoothness, has
triggered people's interest since it is more realistic in many optimization
problems shown by both empirical and theoretical evidence. Two recent
works established the $O(\epsilon^{-3})$ sample complexity to obtain
an $O(\epsilon)$-stationary point. However, both require a large
batch size on the order of $\mathrm{ploy}(\epsilon^{-1})$, which
is not only computationally burdensome but also unsuitable for streaming
applications. Additionally, these existing convergence bounds are
established only for the expected rate, which is inadequate as they
do not supply a useful performance guarantee on a single run. In this
work, we solve the prior two problems simultaneously by revisiting
a simple variant of the STORM algorithm. Specifically, under the $(L_{0},L_{1})$-smoothness
and affine-type noises, we establish the first near-optimal $O(\log(1/(\delta\epsilon))\epsilon^{-3})$
high-probability sample complexity where $\delta\in(0,1)$ is the
failure probability. Besides, for the same algorithm, we also recover
the optimal $O(\epsilon^{-3})$ sample complexity for the expected
convergence with improved dependence on the problem-dependent parameter.
More importantly, our convergence results only require a constant
batch size in contrast to the previous works.
\end{abstract}

\section{Introduction}

In this paper, we consider the stochastic optimization problem of
the form:

\begin{equation}
\min_{x\in\R^{d}}F(x)=\E_{\Xi\sim\domxi}\left[f(x,\Xi)\right],\label{eq:problem}
\end{equation}
where both $F(x)$ and $f(x,\Xi)$ are not necessarily convex and
$\Xi$ is drawn from a possibly unknown probability distribution $\domxi$.
Problem (\ref{eq:problem}) has attracted significant attention from
the optimization community in recent years as many modern machine
learning problems can be cast in such a form.

One famous method for solving Problem (\ref{eq:problem}) is the classic
stochastic gradient descent (SGD) algorithm \citep{robbins1951stochastic},
which is easy to implement and enjoys empirical success. For the theoretical
justification, it is well-known that, under the standard $L$-smooth
condition (i.e., the gradient of $F(x)$ is $L$-Lipschitz) and the
finite variance assumption, SGD guarantees an $O(T^{-\frac{1}{4}})$\footnote{This is equivalent to the $O(\epsilon^{-4})$ sample complexity to
obtain an $O(\epsilon)$-stationary point. In the remainder of this
paper, for the algorithm only requiring at most the constant order
batch size like SGD here, we will also use the rate with respect to
$T$ to describe the corresponding convergence behavior due to it
being interchangeable with the sample complexity.} convergence rate after running $T$ iterations for finding the stationary
point, in other words, minimizing the norm of gradients. The $O(T^{-\frac{1}{4}})$
rate is known to be optimal \citep{arjevani2023lower} without further
assumptions. An important line of work to improve the performance
of algorithms for solving Problem (\ref{eq:problem}) is to add the
variance reduction technique, which was originally proposed to speed
up the convergence rate for convex stochastic problems when the objective
function is considered in the finite sum form. Later, people aware
that such a technique can also accelerate the convergence of algorithms
for Problem (\ref{eq:problem}) when the additional averaged $L$-smooth
condition (i.e., $\E_{\Xi\sim\domxi}\left[\|\na_{x}f(x,\Xi)-\na_{y}f(y,\Xi)\|^{2}\right]\leq L^{2}\|x-y\|^{2}$)
is imposed. Nowadays, several algorithms integrated variance reduction
have been proposed and shown to match the lower bound $\Omega(T^{-\frac{1}{3}})$
proved in \citet{arjevani2023lower}. 

Starting from \citet{Zhang2020Why}, several works pointed out that
the $L$-smooth condition can be violated in different machine learning
tasks, for example, neural networks and distributionally robust optimization
(DRO) problems. Hence, a generalized smooth condition -- $(L_{0},L_{1})$-smoothness
-- was introduced to better fit modern optimization problems. Under
this new relaxed assumption, people have established $O(T^{-\frac{1}{4}})$
for different methods (e.g., the clipping or normalized SGD algorithm).
Until recently, two new works \citep{reisizadeh2023variance, pmlr-v202-chen23ar}
tried to explore whether the variance reduction technique can be applied
to this harder problem. Surprisingly, the answer is positive. Specifically,
they showed that under mild assumptions, the improved $O(\epsilon^{-3})$
sample complexity is still achievable for Problem (\ref{eq:problem})
satisfying the averaged version of $(L_{0},L_{1})$-smoothness condition.

However, there are still some missing parts in \citet{reisizadeh2023variance}
and \citet{pmlr-v202-chen23ar}. First, they both only provide the
in-expectation property of their algorithms\footnote{Though the results provided in \citet{reisizadeh2023variance} are
presented as the convergence in probability, they are done by a simple
application of Markov's inequality to the expected convergence. Hence,
we count it as convergence in expectation here.}. Whereas a bound in expectation contains little information about
a single run of the algorithm convergence, which means an individual
running can produce a poor-quality result. This is far from satisfactory
since re-training for the modern large-scale optimization problem
can be very expensive and time-consuming. Hence, we also need a high-probability
bound for the convergence rate with the dependence of $O(\log(1/\delta))$
where $\delta$ is the failure probability. Second, the existing bounds
in \citet{reisizadeh2023variance,pmlr-v202-chen23ar} are not adaptive
to the problem-dependent parameter meaning. For example, the bounds
in \citet{reisizadeh2023variance} depend on $O(L_{0}/L_{1})$, which
becomes meaningless when $L_{1}=0$ (i.e., the standard $L$-smooth
case). When the noises are zero, the results in \citet{pmlr-v202-chen23ar}
are still in the order of $O(\epsilon^{-3})$ and can not recover
the standard $O(\epsilon^{-2})$ sample complexity. Besides, both
their algorithms require the batch size to depend on the target accuracy
$\epsilon$. This drawback makes it impossible to implement their
algorithms with streaming data. Moreover, the algorithms considered
in the previous works are in a double-loop style, which is relatively
complicated.

In this work, we close these important gaps mentioned above. To be
more precise, for Problem (\ref{eq:problem}) under the stochastic
version of the $(L_{0},L_{1})$-smoothness condition, we show there
exists a single-loop algorithm enjoying the near-optimal $O((\log(T/\delta)/T)^{\frac{1}{3}})$
high-probability convergence rate with probability at least $1-\delta$
and the optimal $O(T^{-\frac{1}{3}})$ expected convergence rate after
$T$ iterations running. The algorithm only requires the batch size
to be at most some constant and without knowing the target accuracy
$\epsilon$. We can even set the batch size to be $1$ in some cases.
More importantly, both two bounds are adaptive to all the problem-dependent
parameters simultaneously meaning that we can recover the existing
results when considering $L_{1}=0$ or the deterministic case.

\subsection{Our Contributions}

The contributions of our paper are listed as follows:

We establish the first near-optimal high-probability convergence result
under the generalized $(L_{0},L_{1})$-smoothness condition and the
affine-type noise assumption by revisiting an existing algorithm,
Normalized STORM. To be more precise, given $\delta\in(0,1)$, the
convergence rate of the algorithm is in the order of $O((\log(T/\delta)/T)^{\frac{1}{3}})$
with probability at least $1-\delta$ where $T$ is the time horizon.
Additionally, the rate is adaptive to all the problem-dependent parameters
at the same time. In contrast, as far as we know, all the existing
high-probability bounds for non-convex stochastic problems with variance
reduction only provide the analysis under the standard smooth condition\textit{.}
Therefore, we believe that our work is an important step to help people
understand the high-probability convergence behavior for optimizing
functions satisfying the generalized smoothness condition.

For the same algorithm, we also prove an expected convergence result
in the optimal rate $O(T^{-\frac{1}{3}})$. Remarkably, our expected
convergence theory reveals that stochastic optimization is as easy
as the deterministic problem in a certain regime. More precisely,
we prove that an improved $O(T^{-\frac{1}{2}})$ expected convergence
rate is achievable for stochastic optimization even under the $(L_{0},L_{1})$-smoothness
condition. Moreover, the expected convergence result is adaptive to
all the problem-dependent parameters as well.

Notably, both kinds of convergence are established using only a constant
batch size in contrast to the previous results requiring the batch
size depending on the target accuracy $\epsilon$. Indeed, for the
high-probability convergence, the batch size can always be set to
$1$. Additionally, the algorithm investigated by us is in a single-loop
style in comparison to the double-loop algorithm considered in the
prior works.

\subsection{Related Work}

\textbf{Generalized $(L_{0},L_{1})$-smoothness:} The generalized
$(L_{0},L_{1})$-smoothness condition was originally introduced by
\citet{Zhang2020Why} for the twice differentiable functions aiming
to develop a better theoretical understanding for the deep learning
model. Later on, \citet{zhang2020improved} extended the original
definition into a new description to fit the functions only required
to be differentiable. More interestingly, \citet{jin2021non} observed
that certain DRO problems are equivalent to minimizing the generalized
$(L_{0},L_{1})$-smooth functions. From the algorithmic side, for
the clipping or normalized SGD algorithm, \citet{Zhang2020Why,zhang2020improved,jin2021non,zhao2021convergence}
established the $O(\epsilon^{-4})$ sample complexity to reach an
$O(\epsilon)$-stationary point (i.e., $\E\left[\|\na F(x)\|\right]\leq O(\epsilon)$)
for the stochastic problem under the $(L_{0},L_{1})$-smoothness condition
with different assumptions on the noises. \citet{crawshaw2022robustness}
proposed a generalized SignSGD algorithm achieving the $\widetilde{O}(\epsilon^{-4})$
sample complexity with a high probability. \citet{pmlr-v195-faw23a,pmlr-v195-wang23a}
proved that the famous adaptive optimization algorithm, AdaGrad-Norm
\citep{McMahanS10, duchi2011adaptive}, can also converge in the expected
rate $\widetilde{O}(T^{-\frac{1}{4}})$ under this relaxed smooth
condition. Until recently, two works \citep{reisizadeh2023variance, pmlr-v202-chen23ar}
first obtained the improved $O(\epsilon^{-3})$ sample complexity
by applying the variance-reduced algorithm to the function class satisfying
the stochastic extension of $(L_{0},L_{1})$-smoothness condition
(see Section \ref{sec:preliminaries} for a detailed definition).

\textbf{Variance reduction for stochastic optimization: }Variance
reduction technique \citep{roux2012stochastic, johnson2013accelerating, shalev2013stochastic, mairal2013optimization, defazio2014saga}
is introduced to improve the convergence rate for convex stochastic
optimization of the finite sum problem. After lots of intermediate
works (e.g., \citet{allen2017katyusha,nguyen2017sarah}), many algorithms
\citep{lan2019unified, zhou2019direct, song2020variance, liu2022adaptive, carmon2022recapp}
are proved to be near-optimal or optimal under different settings.
For non-convex problems, it is also found that the variance reduction
technique can improve the convergence rate in different situations.
A large number of works \citet{fang2018spider,cutkosky2019momentum,tran2019hybrid,liu2020optimal,li2021page}
have matched the lower bound of $\Omega(T^{-\frac{1}{3}})$ proved
by \citet{arjevani2023lower} under mild assumptions when the problem
is in the form of (\ref{eq:problem}), which improves upon the well-known
speed of $\Theta(T^{-\frac{1}{4}})$ for the vanilla SGD or momentum
SGD. Additionally, \citet{huang2021super,levy2021storm+,liu2022meta}
also achieved the rate $\widetilde{O}(T^{-\frac{1}{3}})$ for the
adaptive algorithm.\textbf{}

\section{Preliminaries\label{sec:preliminaries}}

\textbf{Notations:} $\N$ is the set of natural numbers (excluding
$0$). $\left[d\right]$ denotes the set $\left\{ 1,2,\cdots,d\right\} $
for any integer $d\geq1$. $\|\cdot\|$ stands for the Euclidean norm.

\subsection{$(L_{0},L_{1})$-Smoothness}

In this section, we first provide the formal definition of $(L_{0},L_{1})$-smoothness
and then introduce two basic and useful results.

The concept of the $(L_{0},L_{1})$-smoothness was originally proposed
by \citet{Zhang2020Why} but only for twice differentiable functions
satisfying $\|\na^{2}F(x)\|\leq L_{0}+L_{1}\|\na F(x)\|$ for some
$L_{0},L_{1}\geq0$. Later on, \citet{zhang2020improved} relaxed
the twice differentiable requirement and gave an alternative description
of $(L_{0},L_{1})$-smoothness for the differentiable functions, which
is provided in Definition \ref{def:smooth} as follows:
\begin{defn}
\label{def:smooth}A differentiable function $F(x)$ is said to be
$(L_{0},L_{1})$-smooth if there exist $L_{0},L_{1}\geq0$ such that
for any $x,y\in\R^{d}$ satisfying $\|x-y\|\leq\frac{1}{L_{1}}$,
the following inequality holds
\[
\|\na F(x)-\na F(y)\|\leq(L_{0}+L_{1}\|\na F(x)\|)\|x-y\|.
\]
\end{defn}
Note that Definition \ref{def:smooth} reduces to the widely used
condition of $L$-smoothness when $L_{1}=0$. Therefore, this is a
strictly broader function class including the standard $L$-smooth
functions as subcases. Next, we introduce Lemma \ref{lem:descent-lemma}
that is known as the descent inequality for the $(L_{0},L_{1})$-smooth
functions in previous works \citep{zhang2020improved, jin2021non}.
The proof of Lemma \ref{lem:descent-lemma} is included in Appendix
\ref{sec:app-basic} for completeness.
\begin{lem}
\label{lem:descent-lemma}Suppose $F(x)$ is $(L_{0},L_{1})$-smooth,
then for any $x,y\in\R^{d}$ satisfying\textup{ $\|x-y\|\leq\frac{1}{L_{1}}$},
there is
\[
F(y)\leq F(x)+\langle\na F(x),y-x\rangle+\frac{L_{0}+L_{1}\|\na F(x)\|}{2}\|x-y\|^{2}.
\]
\end{lem}
Lastly, we introduce an important property for $(L_{0},L_{1})$-smooth
functions in Lemma \ref{lem:gradient-norm-bound}. As far as we know,
this result is new, the proof of which is provided in Appendix \ref{sec:app-basic}.
\begin{lem}
\label{lem:gradient-norm-bound}Suppose $F(x)$ is $(L_{0},L_{1})$-smooth
and let $\Delta_{x}\coloneqq F(x)-F_{*}$, then there is
\[
\|\na F(x)\|\leq\sqrt{2L_{0}\Delta_{x}}+2L_{1}\Delta_{x}.
\]
\end{lem}
To understand this property, one can first consider the special case
of $L_{1}=0$. In this situation, Lemma \ref{lem:gradient-norm-bound}
degenerates to $\|\na F(x)\|\leq\sqrt{2L_{0}\Delta_{x}}$, which is
the standard inequality for $L$-smooth functions \citep{nesterov2018lectures, lan2020first}.
Hence, Lemma \ref{lem:gradient-norm-bound} can be viewed as an extension
of the previous result to fit the new $(L_{0},L_{1})$-smoothness
assumption. We would like to emphasize that Lemma \ref{lem:gradient-norm-bound}
plays a central role in helping us obtain the high-probability convergence
bound. For a more detailed discussion, we refer the interested reader
to Section \ref{sec:analysis}.

\subsection{Problem Setup}

We focus on the non-convex stochastic optimization problem for which
the objective function $F:\R^{d}\to\R$ is in the form of $F(x)=\E_{\Xi\sim\domxi}\left[f(x,\Xi)\right]$
where $\Xi$ obeys a probability distribution $\domxi$. We will omit
the writing of the subscript $\Xi\sim\domxi$ for simplicity in the
remaining paper. Additionally, both $F$ and $f$ are assumed to be
differentiable with respect to $x$. $\na f(x,\Xi)$ represents $\na_{x}f(x,\Xi)$
for convenience. Our analysis relies on the following assumptions:

\textbf{1. Finite lower bound:} $F_{*}\coloneqq\inf_{x\in\R^{d}}F(x)>-\infty$.

\textbf{2. Unbiased gradients:} $\na F(x)=\E\left[\na f(x,\Xi)\mid x\right]$,
$\forall x\in\R^{d}$

\textbf{3A. Almost sure $(L_{0}$, $L_{1})$-smoothness:} $\exists L_{0},L_{1}\geq0$
such that $\|\na f(x,\Xi)-\na f(y,\Xi)\|\leq(L_{0}+L_{1}\|\na F(x)\|)\|x-y\|,\forall x,y\in\R^{d}$
satisfying $\|x-y\|\leq\frac{1}{L_{1}}$ almost surely.

\textbf{3B. Expected $(L_{0}$, $L_{1})$-smoothness:} $\exists L_{0},L_{1}\geq0$
such that $\E\left[\|\na f(x,\Xi)-\na f(y,\Xi)\|^{2}\mid x,y\right]\leq(L_{0}+L_{1}\|\na F(x)\|)^{2}\|x-y\|^{2},\forall x,y\in\R^{d}$
satisfying $\|x-y\|\leq\frac{1}{L_{1}}$ almost surely.

\textbf{4A. Almost sure $(\sigma_{0},\sigma_{1})$-affine noises:}
$\exists\sigma_{0},\sigma_{1}\geq0$ such that $\|\na f(x,\Xi)-\na F(x)\|\leq\sigma_{0}+\sigma_{1}\|\na F(x)\|$,
$\forall x\in\R^{d}$ almost surely.

\textbf{4B. Expected $(\sigma_{0},\sigma_{1})$-affine noises:} $\exists\sigma_{0},\sigma_{1}\geq0$
such that $\E\left[\|\na f(x,\Xi)-\na F(x)\|^{2}\mid x\right]\leq\sigma_{0}^{2}+\sigma_{1}^{2}\|\na F(x)\|^{2}$,
$\forall x\in\R^{d}$.

Assumptions 1 and 2 are standard and widely used in the related literature
on stochastic optimization problems. Assumptions 3A and 3B are two
variants of the $(L_{0},L_{1})$-smoothness to fit the stochastic
programming and both of which imply that $F(x)$ itself is $(L_{0},L_{1})$-smooth.
Note that Assumption 3B can be viewed as a generalization of averaged
$L$-smooth \citep{arjevani2023lower} and has been used in the previous
works like \citet{reisizadeh2023variance}. Assumptions 4A and 4B
are known as affine-type noises \citep{bottou2018optimization}. The
former is a weaker version of the latter but can help us derive the
high-probability convergence result.

Our high-probability analysis is based on the following technical
tool. Compared with the well-known Hoeffding's inequality \citep{409cf137-dbb5-3eb1-8cfe-0743c3dc925f}
for bounded scaled martingale difference sequence, Lemma \ref{lem:concentration}
is a generalized dimension-free result in the Hilbert Space. A similar
result was proved by \citet{10.1214/aop/1176988477} before. For completeness,
the proof of Lemma \ref{lem:concentration} is included in Section
\ref{sec:app-basic} in the appendix.
\begin{lem}
\label{lem:concentration}Suppose $X_{t\in\left[T\right]}$ is a martingale
difference sequence adapted to the filtration $\F_{t\in\left[T\right]}$
in a Hilbert Space satisfying $\|X_{t}\|\leq R_{t},\forall t\in\left[T\right]$
for some constant $R_{t}\geq0$ almost surely. Then, for any given
$\delta\in(0,1)$, with probability at least $1-\delta$ , there is
\[
\left\Vert \sum_{s=1}^{t}X_{s}\right\Vert \leq4\sqrt{\log\frac{2}{\delta}\sum_{s=1}^{T}R_{s}^{2}},\forall t\in\left[T\right].
\]
\end{lem}

\section{Algorithm and Results\label{sec:algo}}

In this section, we present the Normalized STORM algorithm and provide
its convergence guarantee under the generalized smoothness condition
both in high probability and in expectation.

\subsection{Normalized STORM}

\begin{algorithm}[H]
\caption{Normalized STORM\label{alg:STORM}}

\textbf{Input}: Initial point $x_{1}\in\R^{d}$, batch size $k\leq K\in\N$,
momentum parameter $\beta\in[0,1)$, step size $\eta>0$, time horizon
$T\in\N$

\textbf{for} $t=1$ \textbf{to} $T$ \textbf{do}

$\quad$Draw independent samples $K_{t}=\left\{ \Xi_{t}^{i}\sim\domxi,i\in\left[K\right]\right\} $

$\quad$$\na f(x_{t},K_{t})\coloneqq\frac{1}{K}\sum_{i=1}^{K}\na f(x_{t},\Xi_{t}^{i})$

$\quad$$\na f(x_{t},k_{t})\coloneqq\frac{1}{k}\sum_{i=1}^{k}\na f(x_{t},\Xi_{t}^{i})$

$\quad$$\na f(x_{t-1},k_{t})\coloneqq\frac{1}{k}\sum_{i=1}^{k}\na f(x_{t-1},\Xi_{t}^{i})$

$\quad$$m_{t}=\beta m_{t-1}+(1-\beta)\na f(x_{t},K_{t})+\mathds{1}_{t\geq2}\beta(\na f(x_{t},k_{t})-\na f(x_{t-1},k_{t}))$
where $m_{0}\coloneqq\na f(x_{1},K_{1})$

$\quad$$x_{t+1}=x_{t}-\eta\frac{m_{t}}{\|m_{t}\|}$

\textbf{end for}
\end{algorithm}

The algorithm, Normalized STORM, is shown in Algorithm \ref{alg:STORM},
which is a simple variant of the original STORM algorithm \citep{cutkosky2019momentum}.
Algorithm \ref{alg:STORM} also appeared in \citet{cutkosky2022lecture}
but without considering using the batch of samples. However, we remark
that employing a batch size $K$ being potentially larger than $1$
is the key to establishing the convergence theory of Algorithm \ref{alg:STORM}
under the affine-type noises. We also would like to emphasize that
Algorithm \ref{alg:STORM} is a simple single-loop style algorithm
in contrast to the double-loop variance-reduced algorithm used in
the previous related works \citep{reisizadeh2023variance, pmlr-v202-chen23ar}.

We briefly talk about why the STORM like algorithm can achieve variance
reduction here. For a detailed explanation, the reader could refer
to \citet{cutkosky2019momentum,cutkosky2022lecture}. Let us keep
the batch size $K=k=1$ and assume $t\geq2$ in the following discussion
for simplicity. In this case, one can see the STORM template incorporates
momentum and variance reduction as follows:
\begin{align}
m_{t}= & \underbrace{\beta m_{t-1}+(1-\beta)\na f(x_{t},\Xi_{t})}_{(i)}\nonumber \\
 & +\underbrace{\beta(\na f(x_{t},\Xi_{t})-\na f(x_{t-1},\Xi_{t}))}_{(ii)}.\label{eq:storm}
\end{align}
As shown in (\ref{eq:storm}), the gradient estimator $m_{t}$ can
be viewed as a combination of $(i)$ and $(ii)$. Part $(i)$ is the
same as the gradient estimator in the algorithm SGD with momentum.
Part $(ii)$ is the variance reduction part playing the key role in
obtaining a better convergence rate. By properly choosing the momentum
parameter $\beta$ and the step size $\eta$, several works \citep{tran2019hybrid, cutkosky2019momentum, liu2020optimal, cutkosky2022lecture}
established the $O(T^{-\frac{1}{3}})$ convergence guarantee in expectation
for the averaged $L$-smooth functions under the finite variance condition,
which matches the lower bound proved in \citet{arjevani2023lower}
and is faster than the well-known rate $O(T^{-\frac{1}{4}})$ of SGD
due to the application of the variance-reduced part.

Finally, we would like to discuss the batch size $K$ and $k$ before
moving to the convergence theory. The reason that we use another batch
size $k\leq K$ is to reduce the computational costs since we do not
only compute the gradient at the point $x_{t}$ but also need to compute
the gradient at the point $x_{t-1}$. As indicated by our theoretical
results, $k$ can be indeed any number not larger than $K$. Even
$k=1$ is always allowed.

\subsection{Convergence Guarantee}

We are now ready to state our main results, Theorem \ref{thm:storm-hp-main}
for convergence in high probability and Theorem \ref{thm:storm-exp-main}
for convergence in expectation. The proofs of these two theorems are
deferred into Section \ref{sec:app-full} in the appendix due to limited
space.
\begin{thm}
\label{thm:storm-hp-main}Suppose Assumptions 1, 2, 3A and 4A hold
and let $\Delta_{1}=F(x_{1})-F_{*}$. If $K\geq1$ and $k\in\left[K\right]$
is chosen arbitrarily, then for any given $T\in\N$ and $\delta\in(0,1)$,
under properly picked $\beta$ and $\eta$, Algorithm \ref{alg:STORM}
guarantees that with probability at least $1-\delta$,
\begin{align*}
 & \min_{t\in\left[T\right]}\|\na F(x_{t})\|\leq\\
\widetilde{O} & \left(\frac{\sigma_{0}+\sigma_{1}\|\na F(x_{1})\|}{T}+\frac{\sqrt{\Delta_{1}L_{0}}+\Delta_{1}L_{1}}{\sqrt{T}}\right.\\
 & \left.+\sqrt[3]{\frac{(\sigma_{0}+\sigma_{1}\|\na F(x_{1})\|)(\sigma_{0}+\sigma_{1}(\sqrt{\Delta_{1}L_{0}}+\Delta_{1}L_{1}))^{2}}{T}}\right.\\
 & \left.+\sqrt[3]{\frac{(\sigma_{0}+\sigma_{1}(\sqrt{\Delta_{1}L_{0}}+\Delta_{1}L_{1}))(\sqrt{\Delta_{1}L_{0}}+\Delta_{1}L_{1})^{2}}{T}}\right).
\end{align*}
\end{thm}
We remark that the $\widetilde{O}$-notation only hides the factor
$\log(T/\delta)$. The explicit dependence on $\log(T/\delta)$ and
the precise definitions of $\beta$ and $\eta$ are provided in Theorem
\ref{thm:storm-hp-full} in Appendix \ref{sec:app-full}. To the best
of our knowledge, Theorem \ref{thm:storm-hp-main} is the first high-probability
bound of the variance-reduced algorithm for non-convex stochastic
optimization problems attaining the near-optimal $\widetilde{O}(T^{-\frac{1}{3}})$
rate under the generalized $(L_{0},L_{1})$-smooth condition and the
affine-type noise assumption. We remark that \citet{reisizadeh2023variance}
also provided a convergence in probability result. However, that bound
is obtained by simply applying Markov's inequality to the expected
convergence bound. Hence, the dependence on the failure probability
$\delta$ is in the order of $O(\mathrm{poly}(1/\delta))$, which
is far from the optimal $O(\mathrm{polylog}(1/\delta))$ in Theorem
\ref{thm:storm-hp-main}. We will give a more detailed comparison
to the rate in \citet{reisizadeh2023variance} later when presenting
our expected convergence bound.

There are some advantages we would like to emphasize in this high-probability
result. First, the batch size $K$ and $k$ can be chosen arbitrarily
meaning that we can even set $K=k=1$ to get rid of extra computational
costs. This benefit ensures Algorithm \ref{alg:STORM} can work even
with the streaming data. Next, our convergence rate is adaptive to
the noise parameters $\sigma_{0}$ and $\sigma_{1}$. In other words,
our rate recovers the optimal $O(T^{-\frac{1}{2}})$ rate in the deterministic
case (i.e., $\sigma_{0}=\sigma_{1}=0$).

With Theorem \ref{thm:storm-hp-main}, we can obtain the following
near-optimal sample complexity for Algorithm \ref{alg:STORM}. Again,
note that we can always choose $K=k=1$.
\begin{cor}
Under the same conditions as Theorem \ref{thm:storm-hp-main}, if
we take $K=k$, the number of iterations $T_{\epsilon}$ used to obtain
an $O(\epsilon)$-stationary point is at most
\begin{align*}
\widetilde{O} & \left(\frac{\sigma_{0}+\sigma_{1}\|\na F(x_{1})\|}{\epsilon}+\frac{\Delta_{1}L_{0}+\Delta_{1}^{2}L_{1}^{2}}{\epsilon^{2}}\right.\\
 & \left.+\frac{(\sigma_{0}+\sigma_{1}\|\na F(x_{1})\|)(\sigma_{0}+\sigma_{1}(\sqrt{\Delta_{1}L_{0}}+\Delta_{1}L_{1}))^{2}}{\epsilon^{3}}\right.\\
 & \left.+\frac{(\sigma_{0}+\sigma_{1}(\sqrt{\Delta_{1}L_{0}}+\Delta_{1}L_{1}))(\sqrt{\Delta_{1}L_{0}}+\Delta_{1}L_{1})^{2}}{\epsilon^{3}}\right).
\end{align*}
The number of samples $KT_{\epsilon}$ we need is at most
\begin{align*}
\widetilde{O} & \left(\frac{(\sigma_{0}+\sigma_{1}\|\na F(x_{1})\|)K}{\epsilon}+\frac{(\Delta_{1}L_{0}+\Delta_{1}^{2}L_{1}^{2})K}{\epsilon^{2}}\right.\\
 & \left.+\frac{(\sigma_{0}+\sigma_{1}\|\na F(x_{1})\|)(\sigma_{0}+\sigma_{1}(\sqrt{\Delta_{1}L_{0}}+\Delta_{1}L_{1}))^{2}K}{\epsilon^{3}}\right.\\
 & \left.+\frac{(\sigma_{0}+\sigma_{1}(\sqrt{\Delta_{1}L_{0}}+\Delta_{1}L_{1}))(\sqrt{\Delta_{1}L_{0}}+\Delta_{1}L_{1})^{2}K}{\epsilon^{3}}\right).
\end{align*}
In particular, we can always set the batch size to be $K=k=1$.
\end{cor}
Now let us move to the expected convergence rate as shown in Theorem
\ref{thm:storm-exp-main}. The full version of expected convergence,
Theorem \ref{thm:storm-exp-full} (including the definitions of $\beta$
and $\eta$) and its proof are provided in Section \ref{sec:app-full}
in the appendix.
\begin{thm}
\label{thm:storm-exp-main}Suppose Assumptions 1, 2, 3B and 4B hold
and let $\Delta_{1}=F(x_{1})-F_{*}$. If $K\geq\max\left\{ \left\lceil 64\sigma_{1}^{2}\right\rceil ,1\right\} $
and $k\in\left[K\right]$ is chosen arbitrarily, then for any given
$T\in\N$, under properly picked $\beta$ and $\eta$, Algorithm \ref{alg:STORM}
guarantees that
\begin{align*}
 & \min_{t\in\left[T\right]}\E\left[\|\na F(x_{t})\|\right]\leq\\
O & \left(\frac{\Delta_{1}L_{1}+(\sigma_{0}+\sigma_{1}\|\na F(x_{1})\|)/\sqrt{K}}{T}+\sqrt{\frac{\Delta_{1}L_{0}}{T}}\right.\\
 & \left.+\sqrt[3]{\frac{\sigma_{0}\Delta_{1}L_{0}}{\sqrt{kK}T}+\frac{\sigma_{0}^{2}(\sigma_{0}+\sigma_{1}\|\na F(x_{1})\|)}{K^{3/2}T}+\frac{\sigma_{0}^{2}\Delta_{1}L_{1}}{\sqrt{k}KT}}\right).
\end{align*}
\end{thm}
Theorem \ref{thm:storm-exp-main} is optimal as it attains the lower
bound rate $\Omega(T^{-\frac{1}{3}})$ for the averaged $L$-smooth
functions under the finite variance assumption \citep{arjevani2023lower},
which is a subclass of the functions satisfying Assumptions 3B and
4B. Compared with the high-probability bound, the extra $O(\log T)$
factor is removed in Theorem \ref{thm:storm-exp-main}. However, unlike
Theorem \ref{thm:storm-hp-main}, the batch size $K$ now is at least
$\max\left\{ \left\lceil 64\sigma_{1}^{2}\right\rceil ,1\right\} $.
Another interesting observation is that the expected rate can be improved
to $O(T^{-\frac{1}{2}})$ even if only $\sigma_{0}$ is set to be
$0$. This means that the stochastic problem under the expected $(L_{0},L_{1})$-smoothness
condition and the assumption of $\E\left[\|\na f(x,\Xi)-\na F(x)\|^{2}\mid x\right]\leq\sigma_{1}^{2}\|\na F(x)\|^{2}$
is as easy as deterministic optimization.

Besides, we would like to compare Theorem \ref{thm:storm-exp-main}
with two previous related works \citep{reisizadeh2023variance, pmlr-v202-chen23ar}
that proved the expected convergence under a similar setting. First,
both of their results are based on another variance-reduced algorithm,
SPIDER \citep{fang2018spider}, which is a different double-loop framework
from Algorithm \ref{alg:STORM}. Second, we emphasize that they both
require a large batch size depending on the target accuracy $\epsilon$.
In contrast, the batch size in Theorem \ref{thm:storm-exp-main} only
needs to exceed a constant threshold $\max\left\{ \left\lceil 64\sigma_{1}^{2}\right\rceil ,1\right\} $,
which can be even reduced to $1$ when $\sigma_{1}\leq1/8$. Besides,
we note that the result in \citet{reisizadeh2023variance} can not
be adaptive to $L_{1}$ due to the dependence of $O(L_{0}/L_{1})$
in their bound. In other words, their bound becomes meaningless for
the classic $L$-smooth case (i.e., when $L_{1}=0$). In comparison,
our bound still holds in the case of $L_{1}=0$. Compared with \citet{pmlr-v202-chen23ar},
our bound is better adaptive to the noise parameter. As mentioned
above, the rate can be improved to $O(T^{-\frac{1}{2}})$ when $\sigma_{0}=0$,
which leads to a better $O(\epsilon^{-2})$ sample complexity. However,
the bounds in \citet{pmlr-v202-chen23ar} do not have this advantage. 

Finally, let us convert Theorem \ref{thm:storm-exp-main} into the
following optimal sample complexity.
\begin{cor}
Under the same conditions as Theorem \ref{thm:storm-exp-main}, if
we take $K=k$, the number of iterations $T_{\epsilon}$ used to obtain
an $O(\epsilon)$-stationary point is at most
\begin{align*}
O & \left(\frac{\Delta_{1}L_{1}+(\sigma_{0}+\sigma_{1}\|\na F(x_{1})\|)/\sqrt{K}}{\epsilon}+\frac{\Delta_{1}L_{0}}{\epsilon^{2}}\right.\\
 & \left.+\left(\frac{\Delta_{1}L_{0}}{K}+\frac{\sigma_{0}^{2}+\sigma_{0}\sigma_{1}\|\na F(x_{1})\|+\sigma_{0}\Delta_{1}L_{1}}{K^{3/2}}\right)\frac{\sigma_{0}}{\epsilon^{3}}\right).
\end{align*}
The number of samples $KT_{\epsilon}$ we need is at most
\begin{align*}
O & \left(\frac{\Delta_{1}L_{1}K+(\sigma_{0}+\sigma_{1}\|\na F(x_{1})\|)\sqrt{K}}{\epsilon}+\frac{\Delta_{1}L_{0}K}{\epsilon^{2}}\right.\\
 & \left.+\left(\Delta_{1}L_{0}+\frac{\sigma_{0}^{2}+\sigma_{0}\sigma_{1}\|\na F(x_{1})\|+\sigma_{0}\Delta_{1}L_{1}}{\sqrt{K}}\right)\frac{\sigma_{0}}{\epsilon^{3}}\right).
\end{align*}
In particular, we can always take the constant batch size $K=k=\max\left\{ \left\lceil 64\sigma_{1}^{2}\right\rceil ,1\right\} $.
Notably, if $\sigma_{1}\leq\frac{1}{8}$, the batch size will reduce
to $K=k=1$.
\end{cor}

\section{Theoretical Analysis\label{sec:analysis}}

In this section, we provide the ideas in the analysis and state some
important lemmas used in the proof. Due to the space limitation, the
proofs of all lemmas presented in this section are deferred into Section
\ref{sec:app-analysis} in the appendix.

To proceed with the following analysis, we introduce some notations
for convenience
\begin{align*}
\Delta_{t\in\left[T\right]} & =F(x_{t})-F_{*};\\
\epsilon_{t\in\left\{ 0\right\} \cup\left[T\right]} & =\begin{cases}
\na f(x_{1},K_{1})-\na F(x_{1}) & t=0\\
m_{t}-\na F(x_{t}) & t\in\left[T\right]
\end{cases};\\
Z_{t\in\left[T\right]} & =\mathds{1}_{t\geq2}\left(\na f(x_{t},k_{t})-\na f(x_{t-1},k_{t})\right.\\
 & \left.\qquad\qquad-\na F(x_{t})+\na F(x_{t-1})\right);\\
\xi_{t\in\left[T\right]} & =\na f(x_{t},K_{t})-\na F(x_{t}).
\end{align*}
We define $\F_{t}$ being the natural filtration generated by $\left\{ K_{s}=\left\{ \Xi_{s}^{i}:i\in\left[K\right]\right\} ,\forall s\in\left[t\right]\right\} $.
Note that $x_{t}$ is $\F_{t-1}$ measurable, $\xi_{t}$ and $Z_{t}$
are both adapted to $\F_{t}$.

We first introduce the following anytime descent inequality as a starting
point in the whole proof. Note that the requirement $\eta\leq\frac{1}{L_{1}}$
implies $\|x_{t+1}-x_{t}\|=\eta\leq\frac{1}{L_{1}}$ to make sure
Lemma \ref{lem:descent-lemma} can be applied to $x_{t+1}$ and $x_{t}$.
\begin{lem}
\label{lem:descent-lemma-any-time}Under Assumptions 1-3 (either 3A
or 3B), if $\eta\leq\frac{1}{L_{1}}$, then for any $t\in\left\{ 0\right\} \cup\left[T\right]$,
there is
\begin{align}
 & \Delta_{t+1}+\sum_{s=1}^{t}\eta\|\na F(x_{s})\|\nonumber \\
\leq & \Delta_{1}+\frac{\eta^{2}tL_{0}}{2}+\sum_{s=1}^{t}2\eta\|\epsilon_{s}\|+\frac{\eta^{2}L_{1}}{2}\|\na F(x_{s})\|.\label{eq:any-time}
\end{align}
\end{lem}
Naturally, the major task is to upper bound the term $\|\epsilon_{t}\|$
both in high probability and in expectation due to Lemma \ref{lem:descent-lemma-any-time}.
To do so, we first need to rewrite $\epsilon_{t}$ in a tractable
way as shown in Lemma \ref{lem:decomposition} for the latter calculation.
This representation also appeared in \citet{cutkosky2022lecture,pmlr-v195-liu23c}
before.
\begin{lem}
\label{lem:decomposition}For any $t\in\left[T\right]$, there is
\[
\epsilon_{t}=\beta^{t}\epsilon_{0}+\beta\sum_{s=1}^{t}\beta^{t-s}Z_{s}+(1-\beta)\sum_{s=1}^{t}\beta^{t-s}\xi_{s}.
\]
\end{lem}
With Lemma \ref{lem:descent-lemma-any-time}, we immediately have
\[
\|\epsilon_{t}\|\leq\beta^{t}\|\epsilon_{0}\|+\beta\left\Vert \sum_{s=1}^{t}\beta^{t-s}Z_{s}\right\Vert +(1-\beta)\left\Vert \sum_{s=1}^{t}\beta^{t-s}\xi_{s}\right\Vert .
\]
Note that $\|\epsilon_{0}\|$ (resp., $\E\left[\|\epsilon_{0}\|\right]$)
can be upper bounded by Assumption 4A (resp., Assumption 4B). Hence,
we only need to consider the remaining two terms. In the next two
sections, we will describe the core ideas on how to derive high-probability
or expected bounds for them.

\subsection{Towards High-Probability Convergence}

In this section, we describe the hard parts of the high-probability
proof and introduce our ideas on how to solve the issues. 

As noted above, from the representation of $\epsilon_{t}$ in Lemma
\ref{lem:decomposition}, the major task is to bound $\|\sum_{s=1}^{t}\beta^{t-s}Z_{s}\|$
and $\|\sum_{s=1}^{t}\beta^{t-s}\xi_{s}\|$ in a high probability
way. An important observation is that for any fixed $t\in\left[T\right]$,
both $\beta^{t-s}Z_{s}$ and $\beta^{t-s}\xi_{s}$ for $s\in\left[t\right]$
are two martingale difference sequences. Hence, a natural idea is
to apply the existing concentration inequality (e.g., Freedman's inequality
\citep{freedman1975tail}) to obtain a high-probability bound. However,
this can not be done immediately due to the following two reasons:
First, we want to bound the norm of vector-valued martingale difference
sequences instead of the real-valued sequences. If we simply apply
some martingale concentration inequality to every coordinate, there
will be extra dependence on the dimension $d$. Second but more importantly,
most of the existing concentration inequalities require the martingale
to be almost surely uniformly bounded. However, if we use Assumptions
3A and 4A to bound $\|Z_{s}\|$ and $\|\xi_{s}\|$, there will be
$\|Z_{s}\|\leq O(\eta(L_{0}+L_{1}\|\na F(x_{s})\|))$ and $\|\xi_{s}\|\leq O(\sigma_{0}+\sigma_{1}\|\na F(x_{s})\|)$.
But $\|\na F(x_{s})\|$ doesn't adamit a uniform bound for $s\in\left[t\right]$. 

Due to the above two challenges, one can not apply the existing concentration
inequality directly. Here, we introduce a way used in our proof to
overcome these two points, which is inspired by the recent work of
\citet{pmlr-v195-liu23c}. Let us go back to Lemma \ref{lem:descent-lemma-any-time}
and suppose we can find a uniformly high-probability bound (say $\Delta$)
to control the R.H.S. of (\ref{eq:any-time}) for any time $t\in\left[\tau\right]$
where $\tau\in\left[T\right]$ is some fixed time. Then Lemma \ref{lem:descent-lemma-any-time}
immediately implies a simple but important fact that we can bound
$\Delta_{t}$ for any $t\in\left[\tau+1\right]$ (we can replace $\Delta$
by $\Delta\lor\Delta_{1}$ to make sure $\Delta_{1}$ is also be bounded).
Recall that Lemma \ref{lem:gradient-norm-bound} tells us the gradient
norm at any point $x$ can be upper bounded by the corresponding function
value gap $\Delta_{x}$. Hence, $\|\na F(x_{t})\|$ for any $t\in\left[\tau+1\right]$
admits a uniform upper bound in the order of $O(\sqrt{L_{0}\Delta}+L_{1}\Delta)$
with a high probability. Then we may apply some concentration inequality
to control $\|\sum_{s=1}^{\tau+1}\beta^{\tau+1-s}Z_{s}\|$ and $\|\sum_{s=1}^{\tau+1}\beta^{\tau+1-s}\xi_{s}\|$.
Therefore, the R.H.S. of (\ref{eq:any-time}) can be bounded again
for time $\tau+1$. By doing this argument iteratively, we can finally
bound the R.H.S. of (\ref{eq:any-time}) for time $T$.

The above thought experiment helps us resolve the second point. Hence,
the only left issue is to find a proper dimension-free concentration
inequality for vector-valued martingale difference sequences to deal
with the first hard part. Thanks to Lemma \ref{lem:concentration},
this can be done easily.

With the above idea, we introduce the following two events happening
with a high probability. These two lemmas are the most crucial parts
in the whole proof for the high-probability convergence. 
\begin{lem}
\label{lem:hp-bound-Z}Under Assumptions 2 and 3A, given $\delta\in(0,1)$
and $G>0$, for any $t\in\left[T\right]$, there is
\[
\Pr\left[a_{t}(G)\right]\geq1-\frac{\delta}{2T},
\]
where $a_{t}(G)$ is the event defined as $a_{t}(G)\coloneqq\left\{ \left\Vert \sum_{s=1}^{t}\beta^{t-s}Z_{s}\chi_{s}(G)\right\Vert \leq8\eta(L_{0}+L_{1}G)\sqrt{\frac{\log\frac{4T}{\delta}}{1-\beta}}\right\} $
and $\chi_{s}(G)\coloneqq\mathds{1}\left[\|\na F(x_{s})\|\leq G\right]$
is the indicator random variable.
\end{lem}
\begin{lem}
\label{lem:hp-bound-xi}Under Assumptions 2 and 4A, given $\delta\in(0,1)$
and $G>0$, for any $t\in\left[T\right]$, there is
\[
\Pr\left[b_{t}(G)\right]\geq1-\frac{\delta}{2T},
\]
where $b_{t}(G)$ is the event defined as $b_{t}(G)\coloneqq\left\{ \left\Vert \sum_{s=1}^{t}\beta^{t-s}\xi_{s}\chi_{s}(G)\right\Vert \leq4(\sigma_{0}+\sigma_{1}G)\sqrt{\frac{\log\frac{4T}{\delta}}{1-\beta}}\right\} $
and $\chi_{s}(G)\coloneqq\mathds{1}\left[\|\na F(x_{s})\|\leq G\right]$
is the indicator random variable.
\end{lem}
In Lemmas \ref{lem:hp-bound-Z} and \ref{lem:hp-bound-xi}, the most
important parameter is the term $G$, which is highly related to the
final bound. Hence, the final thing we need to do is to determine
a proper value for $G$. However, finding such a parameter involves
some tedious calculations, which are deferred into the appendix. After
carefully choosing $G$, we can finally use it to prove the high-probability
convergence bound under the generalized $(L_{0},L_{1})$-smoothness
condition and the affine-type noise assumption, i.e., Theorem \ref{thm:storm-hp-main}.
The reader could refer to Section \ref{sec:app-full} in the appendix
for detailed proofs.

\subsection{Towards In-Expectation Convergence}

Compared with the circuitous ideas used in the high-probability convergence
analysis, the expected convergence can be done in a relatively direct
way. As mentioned, the left work is to deal with the two terms, $\E\left[\|\sum_{s=1}^{t}\beta^{t-s}Z_{s}\|\right]$
and $\E\left[\|\sum_{s=1}^{t}\beta^{t-s}\xi_{s}\|\right]$.

First, let us bound $\E\left[\|\sum_{s=1}^{t}\beta^{t-s}Z_{s}\|\right]$.
As mentioned above, $\beta^{t-s}Z_{s},\forall s\in\left[t\right]$
is a martingale difference sequence. A natural idea is to apply the
following argument $\E\left[\|\sum_{s=1}^{t}\beta^{t-s}Z_{s}\|\right]\leq\sqrt{\E\left[\|\sum_{s=1}^{t}\beta^{t-s}Z_{s}\|^{2}\right]}=\sqrt{\sum_{s=1}^{t}\beta^{2t-2s}\E\left[\|Z_{s}\|^{2}\right]}.$
Then we can bound $\E\left[\|Z_{s}\|^{2}\right]\leq O(\eta^{2}(L_{0}^{2}+L_{1}^{2}\E\left[\|\na F(x_{s})\|^{2}\right]))$
by Assumption 3B. However, this will lead to the term $O(\sum_{s=1}^{t}\beta^{t-s}\sqrt{\E\left[\|\na F(x_{s})\|^{2}\right]})$,
which can not be canceled by the term $\E\left[\|\na F(x_{s})\|\right]$
appeared in the L.H.S. of (\ref{eq:any-time}) after taking expectations.

Hence, we need a more careful strategy. It turns out that applying
an argument of conditional expectation recursively rather than taking
expectations once will lead us to the correct inequality as presented
in the following Lemma (\ref{lem:exp-bound-Z}). With a properly designed
step size $\eta$, we can finally eliminate the effect of the redundant
term $\sum_{s=1}^{t}\sqrt{\frac{2}{k}}\eta L_{1}\beta^{t-s}\E\left[\|\na F(x_{s})\|\right]$.
\begin{lem}
\label{lem:exp-bound-Z}Under Assumptions 2 and 3B, for any $t\in\left[T\right]$,
there is
\begin{align*}
\E\left[\left\Vert \sum_{s=1}^{t}\beta^{t-s}Z_{s}\right\Vert \right]\leq & \frac{\sqrt{2}\eta L_{0}}{\sqrt{k(1-\beta)}}\\
 & +\sum_{s=1}^{t}\sqrt{\frac{2}{k}}\eta L_{1}\beta^{t-s}\E\left[\|\na F(x_{s})\|\right].
\end{align*}
\end{lem}
Now let us consider the term $\E\left[\|\sum_{s=1}^{t}\beta^{t-s}\xi_{s}\|\right]$.
By a similar idea used in proof in Lemma \ref{lem:exp-bound-Z}, we
can obtain Lemma \ref{lem:exp-bound-xi}. We note that a similar inequality
was proved by \citet{jin2021non} before.
\begin{lem}
\label{lem:exp-bound-xi}Under Assumptions 2 and 4B, for any $t\in\left[T\right]$,
there is
\begin{align*}
\E\left[\left\Vert \sum_{s=1}^{t}\beta^{t-s}\xi_{s}\right\Vert \right]\leq & \frac{\sigma_{0}}{\sqrt{K(1-\beta)}}\\
 & +\sum_{s=1}^{t}\frac{\sigma_{1}}{\sqrt{K}}\beta^{t-s}\E\left[\|\na F(x_{s})\|\right].
\end{align*}
\end{lem}
Equipped with Lemmas \ref{lem:exp-bound-Z} and \ref{lem:exp-bound-xi},
the following important inequality for the expected convergence can
be obtained.
\begin{lem}
\label{lem:exp-bound-err}Under Assumptions 2, 3B and 4B, for any
$t\in\left[T\right]$, there is
\begin{align*}
\E\left[\|\epsilon_{t}\|\right]\leq & \beta^{t}\frac{\sigma_{0}+\sigma_{1}\|\na F(x_{1})\|}{\sqrt{K}}+\frac{\sqrt{1-\beta}\sigma_{0}}{\sqrt{K}}+\frac{\sqrt{2}\eta L_{0}}{\sqrt{k(1-\beta)}}\\
 & +\sum_{s=1}^{t}\left(\sqrt{\frac{2}{k}}\eta L_{1}+\frac{(1-\beta)\sigma_{1}}{\sqrt{K}}\right)\beta^{t-s}\E\left[\|\na F(x_{s})\|\right].
\end{align*}
\end{lem}
By applying Lemma \ref{lem:exp-bound-err} to Lemma \ref{lem:descent-lemma-any-time},
we can finally prove Theorem \ref{thm:storm-exp-main} by carefully
choosing the momentum parameter $\beta$ and the step size $\eta$.
We refer the reader to the appendix for a complete proof of the Theorem
\ref{thm:storm-exp-main}.

\section{Conclusion\label{sec:conclu}}

In this work, we revisit a simple variant of the STORM algorithm,
Normalized STORM, and prove it is able to converge under the generalized
$(L_{0},L_{1})$-smoothness condition with the affine-type noise assumption.
Specifically, we establish the first near-optimal high-probability
convergence result attaining the rate of $O((\log(T/\delta)/T)^{\frac{1}{3}})$
after $T$ iterations where $\delta\in(0,1)$ is the failure probability.
Moreover, we also obtain the optimal expected $O(T^{-\frac{1}{3}})$
convergence rate. Both of our bounds are adaptive to the problem-dependent
parameters (e.g., the smooth parameter $L_{1}$ and noise parameter
$\sigma_{0}$ and $\sigma_{1}$) and only require a constant batch
size.

There still remain some limitations in our work. For example, our
results highly depend on the prior knowledge of the parameters. Hence,
it would be interesting and important to design a parameter-free algorithm
that can still achieve the optimal rate but without losing the advantages
mentioned above. Besides, the current high-probability bound suffers
an undesired extra term $O(\log T)$, which we hope can be removed
by a refined argument. We leave these questions as the future direction
and look forward to them being addressed.

\clearpage

\bibliographystyle{abbrvnat}
\bibliography{ref}

\clearpage

\appendix
\onecolumn
\aistatstitle{Supplementary Materials}

\section{Missing Proofs in Section \ref{sec:preliminaries}\label{sec:app-basic}}

In this section, we provide the proofs of lemmas presented in Section
\ref{sec:preliminaries}. Lemma \ref{lem:descent-lemma} is standard.
Lemma \ref{lem:gradient-norm-bound} is new as far as we know and
important for the proof of the high-probability convergence as described
in Section \ref{sec:analysis}. The proof of Lemma \ref{lem:concentration}
here is inspired by \citet{cutkosky2021high,pmlr-v195-liu23c}.

\subsection{Proof of Lemma \ref{lem:descent-lemma}}

\begin{proof}
Because $F$ is differentiable, by applying the fundamental theorem
of calculus to $G(t)\coloneqq F(x+t(y-x))$, we know
\begin{align*}
F(y) & =F(x)+\int_{0}^{1}\langle\na F(x+t(y-x)),y-x\rangle\mathrm{d}t\\
 & =F(x)+\langle\na F(x),y-x\rangle+\int_{0}^{1}\langle\na F(x+t(y-x))-\na F(x),y-x\rangle\mathrm{d}t\\
 & \overset{(a)}{\leq}F(x)+\langle\na F(x),y-x\rangle+\int_{0}^{1}\|\na F(x+t(y-x))-\na F(x)\|\|y-x\|\mathrm{d}t\\
 & \overset{(b)}{\leq}F(x)+\langle\na F(x),y-x\rangle+\int_{0}^{1}(L_{0}+L_{1}\|\na F(x)\|)\|x-y\|^{2}t\mathrm{d}t\\
 & =F(x)+\langle\na F(x),y-x\rangle+\frac{L_{0}+L_{1}\|\na F(x)\|}{2}\|x-y\|^{2},
\end{align*}
where $(a)$ is by Cauchy-Schwarz inequality and $(b)$ is due to
the $(L_{0},L_{1})$-smoothness (Definition \ref{def:smooth}).
\end{proof}

\subsection{Proof of Lemma \ref{lem:gradient-norm-bound}}

\begin{proof}
Let $y=x-\frac{\na F(x)}{L_{0}+L_{1}\|\na F(x)\|}$, we have $\|x-y\|\leq\frac{1}{L_{1}}$.
By Lemma \ref{lem:descent-lemma}, there is
\begin{align*}
F(y) & \leq F(x)+\langle\na F(x),y-x\rangle+\frac{L_{0}+L_{1}\|\na F(x)\|}{2}\|x-y\|^{2}\\
 & =F(x)-\frac{\|\na F(x)\|^{2}}{2(L_{0}+L_{1}\|\na F(x)\|)}\\
\Rightarrow\|\na F(x)\|^{2} & \leq2(L_{0}+L_{1}\|\na F(x)\|)(F(x)-F(y))\\
 & \overset{(a)}{\leq}2(L_{0}+L_{1}\|\na F(x)\|)\Delta_{x},
\end{align*}
where $(a)$ is by $F(y)\geq F_{*}=\inf_{x\in\R^{d}}F(x)$. Note that
\begin{align*}
\|\na F(x)\|^{2} & \leq2(L_{0}+L_{1}\|\na F(x)\|)\Delta_{x}\\
\Leftrightarrow(\|\na F(x)\|-L_{1}\Delta_{x})^{2} & \leq L_{1}^{2}\Delta_{x}^{2}+2L_{0}\Delta_{x}\\
\Rightarrow\|\na F(x)\| & \leq L_{1}\Delta_{x}+\sqrt{L_{1}^{2}\Delta_{x}^{2}+2L_{0}\Delta_{x}}\\
 & \overset{(b)}{\leq}\sqrt{2L_{0}\Delta_{x}}+2L_{1}\Delta_{x},
\end{align*}
where $(b)$ is due to $\sqrt{a+b}\le\sqrt{a}+\sqrt{b}$ for any $a,b\geq0$.
\end{proof}

\subsection{Proof of Lemma \ref{lem:concentration}}

\begin{proof}
By Lemma 10 in \citet{cutkosky2021high}, for any $t\in\left[T\right]$
we have
\[
\left\Vert \sum_{s=1}^{t}X_{s}\right\Vert \leq\left|\sum_{s=1}^{t}M_{s}\right|+\sqrt{\max_{s\in\left[t\right]}\|X_{s}\|^{2}+\sum_{s=1}^{t}\|X_{s}\|^{2}},
\]
where $M_{t}\in\F_{t}$ is a martingale difference sequence satisfying
$\left|M_{t}\right|\leq\|X_{t}\|$ almost surely.

By $\|X_{t}\|\leq R_{t}$ almost surely, there is
\begin{equation}
\left\Vert \sum_{s=1}^{t}X_{s}\right\Vert \leq\left|\sum_{s=1}^{t}M_{s}\right|+\sqrt{\max_{s\in\left[t\right]}R_{s}^{2}+\sum_{s=1}^{t}R_{s}^{2}}\leq\left|\sum_{s=1}^{t}M_{s}\right|+\sqrt{2\sum_{s=1}^{T}R_{s}^{2}}.\label{eq:concentration-1}
\end{equation}
Note that $\left|M_{s}\right|\leq\|X_{s}\|\leq R_{s}$ almost surely,
which implies
\[
\E\left[\exp(\lambda M_{s})\mid\F_{s-1}\right]\leq\exp(\lambda^{2}R_{s}^{2}),\forall\lambda\in\R.
\]
Hence, let $\lambda=\sqrt{\frac{\log(2/\delta)}{\sum_{s=1}^{T}R_{s}^{2}}}$
and define $U_{0}\coloneqq1$ and
\[
U_{t}\coloneqq\exp\left(\sum_{s=1}^{t}\lambda M_{s}-\lambda^{2}R_{s}^{2}\right)\in\F_{t},\forall t\in\left[T\right].
\]
We claim $U_{t}$ is a supermartingale by noticing
\[
\E\left[U_{t}\mid\F_{t-1}\right]=U_{t-1}\E\left[\exp\left(\lambda M_{t}-\lambda^{2}R_{t}^{2}\right)\mid\F_{t-1}\right]\leq U_{t-1}.
\]
Now we define the following stopping time
\[
\tau=\min\left\{ t\in\left[T\right]:U_{t}>\frac{2}{\delta}\right\} 
\]
with $\min\emptyset=\infty$. Then
\begin{align*}
\Pr\left[\exists t\in\left[T\right],U_{t}>\frac{2}{\delta}\right] & =\Pr\left[\tau\leq T\right]\leq\frac{\delta}{2}\E\left[U_{\tau}\mathds{1}\left[\tau\leq T\right]\right]\\
 & =\frac{\delta}{2}\E\left[U_{\tau\land T}\mathds{1}\left[\tau\leq T\right]\right]\leq\frac{\delta}{2}\E\left[U_{\tau\land T}\right]\\
 & \overset{(a)}{=}\frac{\delta}{2}U_{0}=\frac{\delta}{2}
\end{align*}
where $(a)$ is by the optional stopping theorem. Thus, we have
\begin{align*}
 & \Pr\left[\forall t\in\left[T\right],U_{t}\leq\frac{2}{\delta}\right]\geq1-\frac{\delta}{2}\\
\Rightarrow & \Pr\left[\forall t\in\left[T\right],\sum_{s=1}^{t}M_{s}\leq\lambda^{-1}\log\frac{2}{\delta}+\lambda\sum_{s=1}^{t}R_{s}^{2}\right]\geq1-\frac{\delta}{2}\\
\Rightarrow & \Pr\left[\forall t\in\left[T\right],\sum_{s=1}^{t}M_{s}\leq2\sqrt{\log\frac{2}{\delta}\sum_{s=1}^{T}R_{s}^{2}}\right]\geq1-\frac{\delta}{2}.
\end{align*}
By a similar argument, we can obtain
\[
\Pr\left[\forall t\in\left[T\right],\sum_{s=1}^{t}M_{s}\geq-2\sqrt{\log\frac{2}{\delta}\sum_{s=1}^{T}R_{s}^{2}}\right]\geq1-\frac{\delta}{2}.
\]
Combining two cases to get with probability at least $1-\delta$
\begin{equation}
\left|\sum_{s=1}^{t}M_{s}\right|\leq2\sqrt{\log\frac{2}{\delta}\sum_{s=1}^{T}R_{s}^{2}},\forall t\in\left[T\right].\label{eq:concentration-2}
\end{equation}
Finally, plugging (\ref{eq:concentration-2}) into (\ref{eq:concentration-1}),
we have with probability at least $1-\delta$ for any $t\in\left[T\right]$
\[
\left\Vert \sum_{s=1}^{t}X_{s}\right\Vert \leq2\sqrt{\log\frac{2}{\delta}\sum_{s=1}^{T}R_{s}^{2}}+\sqrt{2\sum_{s=1}^{T}R_{s}^{2}}\leq4\sqrt{\log\frac{2}{\delta}\sum_{s=1}^{T}R_{s}^{2}}.
\]
\end{proof}

\section{Full Statements of Main Theorems and Proofs\label{sec:app-full}}

We present the full statements of our two main theorems and provide
their proofs in this section.

\subsection{High-Probability Bound}

In this part, we introduce the high-probability convergence bound.
\begin{thm}
\label{thm:storm-hp-full}Suppose Assumptions 1, 2, 3A and 4A hold
and let $\Delta_{1}=F(x_{1})-F_{*}$. If $K\geq1$ and $k\in\left[K\right]$
is chosen arbitrarily, then for any given $T\in\N$ and $\delta\in(0,1)$,
by taking
\begin{align*}
1-\beta= & \min\left\{ 1,\max\left\{ \left(\frac{\sigma_{0}+\sigma_{1}\|\na F(x_{1})\|}{(\sigma_{0}+\sigma_{1}(\sqrt{\Delta_{1}L_{0}}+\Delta_{1}L_{1}))T\sqrt{\log\frac{4T}{\delta}}}\right)^{\frac{2}{3}},\left(\frac{(\sqrt{\Delta_{1}L_{0}}+\Delta_{1}L_{1})^{2}}{(\sigma_{0}+\sigma_{1}(\sqrt{\Delta_{1}L_{0}}+\Delta_{1}L_{1}))^{2}T\sqrt{\log\frac{4T}{\delta}}}\right)^{\frac{2}{3}}\right\} \right\} ,\\
\eta= & \min\left\{ \sqrt{\frac{\Delta_{1}\sqrt{1-\beta}}{L_{0}T\sqrt{\log\frac{4T}{\delta}}}},\frac{\sqrt{\Delta_{1}}}{\sigma_{1}T\sqrt{L_{0}(1-\beta)\log\frac{4T}{\delta}}},\frac{1}{64\sigma_{1}L_{1}T\sqrt{(1-\beta)\log\frac{4T}{\delta}}},\frac{(1-\beta)^{1/4}}{8\sqrt{2}L_{1}\sqrt{T}(\log\frac{4T}{\delta})^{1/4}}\right\} ,
\end{align*}
Algorithm \ref{alg:STORM} guarantees that with probability at least
$1-\delta$,
\begin{align*}
\min_{t\in\left[T\right]}\|\na F(x_{t})\|\leq & O\left(\frac{\sigma_{0}+\sigma_{1}\|\na F(x_{1})\|}{T}+\frac{(\sqrt{\Delta_{1}L_{0}}+\Delta_{1}L_{1})\log^{\frac{1}{4}}\frac{T}{\delta}}{\sqrt{T}}\right.\\
 & \left.+\sqrt[3]{\frac{(\sigma_{0}+\sigma_{1}\|\na F(x_{1})\|)(\sigma_{0}+\sigma_{1}(\sqrt{\Delta_{1}L_{0}}+\Delta_{1}L_{1}))^{2}\log\frac{T}{\delta}}{T}}\right.\\
 & \left.+\sqrt[3]{\frac{(\sigma_{0}+\sigma_{1}(\sqrt{\Delta_{1}L_{0}}+\Delta_{1}L_{1}))(\sqrt{\Delta_{1}L_{0}}+\Delta_{1}L_{1})^{2}\log\frac{T}{\delta}}{T}}\right).
\end{align*}
\end{thm}
\begin{proof}
Let
\begin{align*}
M & \coloneqq\Delta_{1}+\frac{33\eta^{2}TL_{0}}{2}\sqrt{\frac{\log\frac{4T}{\delta}}{1-\beta}}+\frac{2\eta(\sigma_{0}+\sigma_{1}\|\na F(x_{1})\|)}{1-\beta}+8\eta T\sigma_{0}\sqrt{(1-\beta)\log\frac{4T}{\delta}};\\
N & \coloneqq16\eta^{2}TL_{1}\sqrt{\frac{\log\frac{4T}{\delta}}{1-\beta}}+8\eta T\sigma_{1}\sqrt{(1-\beta)\log\frac{4T}{\delta}};\\
\Delta & \coloneqq4M+8L_{0}N^{2};\\
G & \coloneqq\sqrt{2L_{0}\Delta}+2L_{1}\Delta.
\end{align*}
We first prove that there is
\begin{equation}
M+N\sqrt{2L_{0}\Delta}\leq\frac{\Delta}{2}.\label{eq:storm-hp-ineq}
\end{equation}
Note that this is enough to prove
\[
\sqrt{\Delta}\geq\sqrt{2L_{0}}N+\sqrt{2L_{0}N^{2}+2M}\Leftrightarrow\Delta\geq\left(\sqrt{2L_{0}}N+\sqrt{2L_{0}N^{2}+2M}\right)^{2}.
\]
The last inequality holds due to
\[
\left(\sqrt{2L_{0}}N+\sqrt{2L_{0}N^{2}+2M}\right)^{2}\leq2\left[\left(\sqrt{2L_{0}}N\right)^{2}+2L_{0}N^{2}+2M\right]=\Delta.
\]

Now define the event $e_{t}\coloneqq\left\{ \Delta_{t+1}+\sum_{s=1}^{t}\frac{\eta}{2}\|\na F(x_{s})\|\leq\Delta\right\} ,\forall t\in\left\{ 0\right\} \cup\left[T\right]$.
Besides, we introduce the following three events
\begin{eqnarray*}
E_{\tau}=\cap_{t=0}^{\tau}e_{t}, & A_{\tau}=\cap_{t=1}^{\tau}a_{t}(G), & B_{\tau}=\cap_{t=1}^{\tau}b_{t}(G),
\end{eqnarray*}
where $a_{t}(G)$ and $b_{t}(G)$ are defined in Lemmas \ref{lem:hp-bound-Z}
and \ref{lem:hp-bound-xi} respectively. Our goal is using induction
to prove
\begin{equation}
\Pr\left[G_{\tau}\coloneqq E_{\tau}\cap A_{\tau}\cap B_{\tau}\right]\geq1-\frac{\tau\delta}{T},\forall\tau\in\left\{ 0\right\} \cup\left[T\right].\label{eq:storm-hp-1}
\end{equation}

For $\tau=0$, we know $G_{0}=\left\{ \Delta_{1}\leq\Delta\right\} $
is always true, which means $\Pr\left[G_{0}\right]=1-\frac{0\cdot\delta}{T}$.
Given $\tau\in\left[T\right]$, suppose (\ref{eq:storm-hp-1}) holds
for time $\tau-1$, For time $\tau$, we consider the following event
\[
E_{\tau-1}\cap A_{\tau}\cap B_{\tau}=G_{\tau-1}\cap a_{\tau}(G)\cap b_{\tau}(G)
\]
From Lemmas \ref{lem:hp-bound-Z} and \ref{lem:hp-bound-xi}, we have
\[
\Pr\left[a_{\tau}(G)\right]\geq1-\frac{\delta}{2T},\Pr\left[b_{\tau}(G)\right]\geq1-\frac{\delta}{2T}.
\]
Combining our induction hypothesis (\ref{eq:storm-hp-1}) for time
$\tau-1$, there is
\[
\Pr\left[E_{\tau-1}\cap A_{\tau}\cap B_{\tau}\right]=\Pr\left[G_{\tau-1}\cap a_{\tau}(G)\cap b_{\tau}(G)\right]\geq1-\frac{\tau\delta}{T}.
\]
Now under the event $E_{\tau-1}\cap A_{\tau}\cap B_{\tau}$, we invoke
Lemma \ref{lem:descent-lemma-any-time} for time $\tau$ (this can
be done due to $\eta\leq\frac{(1-\beta)^{1/4}}{8\sqrt{2}L_{1}\sqrt{T}(\log\frac{4T}{\delta})^{1/4}}\leq\frac{1}{L_{1}}$)
to get
\begin{align}
\Delta_{\tau+1}+\sum_{s=1}^{\tau}\eta\|\na F(x_{s})\| & \leq\Delta_{1}+\frac{\eta^{2}\tau L_{0}}{2}+\sum_{s=1}^{\tau}2\eta\|\epsilon_{s}\|+\frac{\eta^{2}L_{1}}{2}\|\na F(x_{s})\|\nonumber \\
 & \overset{(a)}{\leq}\Delta_{1}+\frac{\eta^{2}\tau L_{0}}{2}+\sum_{s=1}^{\tau}2\eta\|\epsilon_{s}\|+\frac{\eta}{2}\|\na F(x_{s})\|\nonumber \\
\Rightarrow\Delta_{\tau+1}+\sum_{s=1}^{\tau}\frac{\eta}{2}\|\na F(x_{s})\| & \leq\Delta_{1}+\frac{\eta^{2}\tau L_{0}}{2}+\sum_{s=1}^{\tau}2\eta\|\epsilon_{s}\|\nonumber \\
 & \overset{(b)}{\leq}\Delta_{1}+\frac{\eta^{2}\tau L_{0}}{2}+\sum_{s=1}^{\tau}2\eta\left(\beta^{s}\|\epsilon_{0}\|+\beta\left\Vert \sum_{t=1}^{s}\beta^{s-t}Z_{t}\right\Vert +(1-\beta)\left\Vert \sum_{t=1}^{s}\beta^{s-t}\xi_{t}\right\Vert \right)\nonumber \\
 & \overset{(c)}{\leq}\Delta_{1}+\frac{\eta^{2}\tau L_{0}}{2}+\frac{2\eta(\sigma_{0}+\sigma_{1}\|\na F(x_{1})\|)}{1-\beta}+\sum_{s=1}^{\tau}2\eta\left(\left\Vert \sum_{t=1}^{s}\beta^{s-t}Z_{t}\right\Vert +(1-\beta)\left\Vert \sum_{t=1}^{s}\beta^{s-t}\xi_{t}\right\Vert \right),\label{eq:storm-hp-2}
\end{align}
where $(a)$ is by $\eta\leq\frac{1}{L_{1}}$, $(b)$ holds due to
Lemma \ref{lem:decomposition}, $(c)$ is true because $\beta\leq1$
and
\[
\|\epsilon_{0}\|=\|\na f(x_{1},K_{1})-\na F(x_{1})\|\leq\frac{1}{K}\sum_{i=1}^{K}\|\na f(x_{1},\Xi_{1}^{i})-\na F(x_{1})\|\leq\sigma_{0}+\sigma_{1}\|\na F(x_{1})\|
\]
from Assumption 4A.

Next, we observe that under the event $E_{\tau-1}$, there is always
\[
\Delta_{t+1}+\sum_{s=1}^{t}\frac{\eta}{2}\|\na F(x_{s})\|\leq\Delta,\forall t\in\left\{ 0\right\} \cup\left[\tau-1\right]\Rightarrow\Delta_{t}\leq\Delta,\forall t\in\left[\tau\right].
\]
Hence, by Lemma \ref{lem:gradient-norm-bound}, we know
\[
\|\na F(x_{t})\|\leq\sqrt{2L_{0}\Delta_{t}}+2L_{1}\Delta_{t}\leq\sqrt{2L_{0}\Delta}+2L_{1}\Delta=G,\forall t\in\left[\tau\right],
\]
which implies
\begin{align*}
\beta^{s-t}Z_{t} & =\beta^{s-t}Z_{t}\mathds{1}\left[\|\na F(x_{t})\|\leq G\right],\forall t\in\left[\tau\right];\\
\beta^{s-t}\xi_{t} & =\beta^{s-t}\xi_{t}\mathds{1}\left[\|\na F(x_{t})\|\leq G\right],\forall t\in\left[\tau\right].
\end{align*}
Thus, under the event $E_{\tau-1}\cap A_{\tau}\cap B_{\tau}$, (\ref{eq:storm-hp-2})
implies
\begin{align*}
 & \Delta_{\tau+1}+\sum_{s=1}^{\tau}\frac{\eta}{2}\|\na F(x_{s})\|\\
\leq & \Delta_{1}+\frac{\eta^{2}\tau L_{0}}{2}+\frac{2\eta(\sigma_{0}+\sigma_{1}\|\na F(x_{1})\|)}{1-\beta}\\
 & +\sum_{s=1}^{\tau}2\eta\left(\left\Vert \sum_{t=1}^{s}\beta^{s-t}Z_{t}\mathds{1}\left[\|\na F(x_{t})\|\leq G\right]\right\Vert +(1-\beta)\left\Vert \sum_{t=1}^{s}\beta^{s-t}\xi_{t}\mathds{1}\left[\|\na F(x_{t})\|\leq G\right]\right\Vert \right)\\
\overset{(d)}{\leq} & \Delta_{1}+\frac{\eta^{2}\tau L_{0}}{2}+\frac{2\eta(\sigma_{0}+\sigma_{1}\|\na F(x_{1})\|)}{1-\beta}+\sum_{s=1}^{\tau}2\eta\left(8\eta(L_{0}+L_{1}G)\sqrt{\frac{\log\frac{4T}{\delta}}{1-\beta}}+4(\sigma_{0}+\sigma_{1}G)\sqrt{(1-\beta)\log\frac{4T}{\delta}}\right)\\
\overset{(e)}{\leq} & \Delta_{1}+\frac{\eta^{2}TL_{0}}{2}+\frac{2\eta(\sigma_{0}+\sigma_{1}\|\na F(x_{1})\|)}{1-\beta}+16\eta^{2}T(L_{0}+L_{1}G)\sqrt{\frac{\log\frac{4T}{\delta}}{1-\beta}}+8\eta T(\sigma_{0}+\sigma_{1}G)\sqrt{(1-\beta)\log\frac{4T}{\delta}}\\
\leq & \Delta_{1}+\frac{33\eta^{2}TL_{0}}{2}\sqrt{\frac{\log\frac{4T}{\delta}}{1-\beta}}+\frac{2\eta(\sigma_{0}+\sigma_{1}\|\na F(x_{1})\|)}{1-\beta}+8\eta T\sigma_{0}\sqrt{(1-\beta)\log\frac{4T}{\delta}}\\
 & +\left(16\eta^{2}TL_{1}\sqrt{\frac{\log\frac{4T}{\delta}}{1-\beta}}+8\eta T\sigma_{1}\sqrt{(1-\beta)\log\frac{4T}{\delta}}\right)G\\
= & M+NG,
\end{align*}
where $(d)$ is by the events $A_{\tau}$ and $B_{\tau}$ happening
and $(e)$ is by $\tau\leq T$. Note that we can bound
\begin{align*}
NG & =N\sqrt{2L_{0}\Delta}+\left[32\eta^{2}TL_{1}^{2}\sqrt{\frac{\log\frac{4T}{\delta}}{1-\beta}}+16\eta T\sigma_{1}L_{1}\sqrt{(1-\beta)\log\frac{4T}{\delta}}\right]\Delta\\
 & \leq N\sqrt{2L_{0}\Delta}+\frac{\Delta}{2},
\end{align*}
where the last inequality holds due to
\[
\eta\leq\begin{cases}
\frac{(1-\beta)^{1/4}}{8\sqrt{2}L_{1}\sqrt{T}(\log\frac{4T}{\delta})^{1/4}} & \Rightarrow32\eta^{2}TL_{1}^{2}\sqrt{\frac{\log\frac{4T}{\delta}}{1-\beta}}\leq\frac{1}{4}\\
\frac{1}{64\sigma_{1}L_{1}T\sqrt{(1-\beta)\log\frac{4T}{\delta}}} & \Rightarrow16\eta T\sigma_{1}L_{1}\sqrt{(1-\beta)\log\frac{4T}{\delta}}\leq\frac{1}{4}
\end{cases}.
\]
So we know
\[
\Delta_{\tau+1}+\sum_{s=1}^{\tau}\frac{\eta}{2}\|\na F(x_{s})\|\leq M+N\sqrt{2L_{0}\Delta}+\frac{\Delta}{2}\leq\Delta,
\]
where the last inequality holds due to (\ref{eq:storm-hp-ineq}).

Hence, under the event $E_{\tau-1}\cap A_{\tau}\cap B_{\tau}$, the
event $e_{\tau}$ happens. This means that $E_{\tau-1}\cap A_{\tau}\cap B_{\tau}\subseteq e_{\tau}$,
which implies
\[
\Pr\left[G_{\tau}\right]=\Pr\left[e_{\tau}\cap E_{\tau-1}\cap A_{\tau}\cap B_{\tau}\right]=\Pr\left[E_{\tau-1}\cap A_{\tau}\cap B_{\tau}\right]\geq1-\frac{\delta\tau}{T}.
\]
Therefore, the induction is completed. Now we know with probability
at least $1-\delta$
\begin{align*}
\Delta_{T+1}+\sum_{t=1}^{T}\frac{\eta}{2}\|\na F(x_{t})\|\leq & \Delta=4M+8L_{0}N^{2}\\
\leq & 4\Delta_{1}+66\eta^{2}TL_{0}\sqrt{\frac{\log\frac{4T}{\delta}}{1-\beta}}+\frac{8\eta(\sigma_{0}+\sigma_{1}\|\na F(x_{1})\|)}{1-\beta}+32\eta T\sigma_{0}\sqrt{(1-\beta)\log\frac{4T}{\delta}}\\
 & +16L_{0}\left(16^{2}\eta^{4}T^{2}L_{1}^{2}\frac{\log\frac{4T}{\delta}}{1-\beta}+8^{2}\eta^{2}T^{2}\sigma_{1}^{2}(1-\beta)\log\frac{4T}{\delta}\right)\\
\overset{(f)}{\leq} & 4\Delta_{1}+98\eta^{2}TL_{0}\sqrt{\frac{\log\frac{4T}{\delta}}{1-\beta}}+1024\eta^{2}T^{2}\sigma_{1}^{2}L_{0}(1-\beta)\log\frac{4T}{\delta}\\
 & +\frac{8\eta(\sigma_{0}+\sigma_{1}\|\na F(x_{1})\|)}{1-\beta}+32\eta T\sigma_{0}\sqrt{(1-\beta)\log\frac{4T}{\delta}}\\
= & O\left(\Delta_{1}+\eta^{2}TL_{0}\sqrt{\frac{\log\frac{4T}{\delta}}{1-\beta}}+\eta^{2}T^{2}\sigma_{1}^{2}L_{0}(1-\beta)\log\frac{4T}{\delta}\right.\\
 & \left.+\frac{\eta(\sigma_{0}+\sigma_{1}\|\na F(x_{1})\|)}{1-\beta}+\eta T\sigma_{0}\sqrt{(1-\beta)\log\frac{4T}{\delta}}\right),
\end{align*}
where $(f)$ is due to $\eta\leq\frac{(1-\beta)^{1/4}}{8\sqrt{2}L_{1}\sqrt{T}(\log\frac{4T}{\delta})^{1/4}}$.
Hence, we have
\begin{align*}
 & \sum_{t=1}^{T}\|\na F(x_{t})\|\\
\leq & O\left(\frac{\Delta_{1}}{\eta}+\eta TL_{0}\sqrt{\frac{\log\frac{4T}{\delta}}{1-\beta}}+\eta T^{2}\sigma_{1}^{2}L_{0}(1-\beta)\log\frac{4T}{\delta}+\frac{\sigma_{0}+\sigma_{1}\|\na F(x_{1})\|}{1-\beta}+T\sigma_{0}\sqrt{(1-\beta)\log\frac{4T}{\delta}}\right)\\
\leq & O\left((\sqrt{\Delta_{1}L_{0}}+\Delta_{1}L_{1})\sqrt{T}\left(\frac{\log\frac{4T}{\delta}}{1-\beta}\right)^{\frac{1}{4}}+\frac{\sigma_{0}+\sigma_{1}\|\na F(x_{1})\|}{1-\beta}+T(\sigma_{0}+\sigma_{1}(\sqrt{\Delta_{1}L_{0}}+\Delta_{1}L_{1}))\sqrt{(1-\beta)\log\frac{4T}{\delta}}\right)\\
\leq & O\left(\sigma_{0}+\sigma_{1}\|\na F(x_{1})\|+(\sqrt{\Delta_{1}L_{0}}+\Delta_{1}L_{1})\sqrt{T}\log^{\frac{1}{4}}\frac{T}{\delta}\right.\\
 & \left.+\left((\sigma_{0}+\sigma_{1}\|\na F(x_{1})\|)(\sigma_{0}+\sigma_{1}(\sqrt{\Delta_{1}L_{0}}+\Delta_{1}L_{1}))^{2}T^{2}\log\frac{T}{\delta}\right)^{\frac{1}{3}}\right.\\
 & \left.+\left((\sigma_{0}+\sigma_{1}(\sqrt{\Delta_{1}L_{0}}+\Delta_{1}L_{1}))(\sqrt{\Delta_{1}L_{0}}+\Delta_{1}L_{1})^{2}T^{2}\log\frac{T}{\delta}\right)^{\frac{1}{3}}\right)
\end{align*}
where the last two inequalities are by plugging in
\begin{align*}
\eta= & \min\left\{ \sqrt{\frac{\Delta_{1}\sqrt{1-\beta}}{L_{0}T\sqrt{\log\frac{4T}{\delta}}}},\frac{\sqrt{\Delta_{1}}}{\sigma_{1}T\sqrt{L_{0}(1-\beta)\log\frac{4T}{\delta}}},\frac{1}{64\sigma_{1}L_{1}T\sqrt{(1-\beta)\log\frac{4T}{\delta}}},\frac{(1-\beta)^{1/4}}{8\sqrt{2}L_{1}\sqrt{T}(\log\frac{4T}{\delta})^{1/4}}\right\} ,\\
1-\beta= & \min\left\{ 1,\max\left\{ \left(\frac{\sigma_{0}+\sigma_{1}\|\na F(x_{1})\|}{(\sigma_{0}+\sigma_{1}(\sqrt{\Delta_{1}L_{0}}+\Delta_{1}L_{1}))T\sqrt{\log\frac{4T}{\delta}}}\right)^{\frac{2}{3}},\left(\frac{(\sqrt{\Delta_{1}L_{0}}+\Delta_{1}L_{1})^{2}}{(\sigma_{0}+\sigma_{1}(\sqrt{\Delta_{1}L_{0}}+\Delta_{1}L_{1}))^{2}T\sqrt{\log\frac{4T}{\delta}}}\right)^{\frac{2}{3}}\right\} \right\} ,
\end{align*}
Finally, we can obtain with probability at least $1-\delta$
\begin{align*}
\min_{t\in\left[T\right]}\|\na F(x_{t})\|= & O\left(\frac{\sigma_{0}+\sigma_{1}\|\na F(x_{1})\|}{T}+\frac{(\sqrt{\Delta_{1}L_{0}}+\Delta_{1}L_{1})\log^{\frac{1}{4}}\frac{T}{\delta}}{\sqrt{T}}\right.\\
 & \left.+\sqrt[3]{\frac{(\sigma_{0}+\sigma_{1}\|\na F(x_{1})\|)(\sigma_{0}+\sigma_{1}(\sqrt{\Delta_{1}L_{0}}+\Delta_{1}L_{1}))^{2}\log\frac{T}{\delta}}{T}}\right.\\
 & \left.+\sqrt[3]{\frac{(\sigma_{0}+\sigma_{1}(\sqrt{\Delta_{1}L_{0}}+\Delta_{1}L_{1}))(\sqrt{\Delta_{1}L_{0}}+\Delta_{1}L_{1})^{2}\log\frac{T}{\delta}}{T}}\right).
\end{align*}
\end{proof}

\subsection{In-Expectaion Rate}

In this section, our goal is to prove the in-expectation convergence
rate.
\begin{thm}
\label{thm:storm-exp-full}Suppose Assumptions 1, 2, 3B and 4B hold
and let $\Delta_{1}=F(x_{1})-F_{*}$. If $K\geq\max\left\{ \left\lceil 64\sigma_{1}^{2}\right\rceil ,1\right\} $
and $k\in\left[K\right]$ is chosen arbitrarily, then for any given
$T\in\N$, by taking
\begin{align*}
1-\beta= & \min\left\{ 1,\max\left\{ \left(\frac{\Delta_{1}L_{0}K}{\sigma_{0}^{2}\sqrt{k}T}\right)^{\frac{2}{3}},\left(\frac{\sigma_{0}+\sigma_{1}\|\na F(x_{1})\|+\Delta_{1}L_{1}\sqrt{K/k}}{\sigma_{0}T}\right)^{\frac{2}{3}}\right\} \right\} ,\\
\eta= & \min\left\{ \sqrt{\frac{\Delta_{1}\min\left\{ \sqrt{k(1-\beta)},1\right\} }{TL_{0}}},\frac{1-\beta}{2(4\sqrt{\frac{2}{k}}+1-\beta)L_{1}}\right\} ,
\end{align*}
Algorithm \ref{alg:STORM} guarantees that
\begin{align*}
\min_{t\in\left[T\right]}\E\left[\|\na F(x_{t})\|\right]\leq & O\left(\frac{\Delta_{1}L_{1}+(\sigma_{0}+\sigma_{1}\|\na F(x_{1})\|)/\sqrt{K}}{T}+\sqrt{\frac{\Delta_{1}L_{0}}{T}}\right.\\
 & \left.+\sqrt[3]{\frac{\sigma_{0}\Delta_{1}L_{0}}{\sqrt{kK}T}+\frac{\sigma_{0}^{2}(\sigma_{0}+\sigma_{1}\|\na F(x_{1})\|)}{K^{3/2}T}+\frac{\sigma_{0}^{2}\Delta_{1}L_{1}}{\sqrt{k}KT}}\right).
\end{align*}
\end{thm}
\begin{proof}
Note that the step size $\eta\leq\frac{1-\beta}{2(4\sqrt{\frac{2}{k}}+1-\beta)L_{1}}\leq\frac{1}{L_{1}}$,
we invoke Lemma \ref{lem:descent-lemma-any-time} for time $T$ to
get
\[
\Delta_{T+1}+\sum_{t=1}^{T}\eta\|\na F(x_{t})\|\leq\Delta_{1}+\frac{\eta^{2}TL_{0}}{2}+\sum_{t=1}^{T}2\eta\|\epsilon_{t}\|+\frac{\eta^{2}L_{1}}{2}\|\na F(x_{t})\|.
\]
Taking expectations on both sides, we obtain
\begin{equation}
\E\left[\Delta_{T+1}\right]+\sum_{t=1}^{T}\eta\E\left[\|\na F(x_{t})\|\right]\leq\Delta_{1}+\frac{\eta^{2}TL_{0}}{2}+\sum_{t=1}^{T}2\eta\E\left[\|\epsilon_{t}\|\right]+\frac{\eta^{2}L_{1}}{2}\E\left[\|\na F(x_{t})\|\right].\label{eq:storm-exp-1}
\end{equation}

By Lemma \ref{lem:exp-bound-err}, we have
\begin{align}
 & \sum_{t=1}^{T}2\eta\E\left[\|\epsilon_{t}\|\right]\nonumber \\
\leq & \sum_{t=1}^{T}2\eta\left(\beta^{t}\frac{\sigma_{0}+\sigma_{1}\|\na F(x_{1})\|}{\sqrt{K}}+\frac{\sqrt{1-\beta}\sigma_{0}}{\sqrt{K}}+\frac{\sqrt{2}\eta L_{0}}{\sqrt{k(1-\beta)}}+\sum_{s=1}^{t}\left(\sqrt{\frac{2}{k}}\eta L_{1}+\frac{(1-\beta)\sigma_{1}}{\sqrt{K}}\right)\beta^{t-s}\E\left[\|\na F(x_{s})\|\right]\right)\nonumber \\
\leq & \frac{2\eta(\sigma_{0}+\sigma_{1}\|\na F(x_{1})\|)}{(1-\beta)\sqrt{K}}+\frac{2\eta\sqrt{1-\beta}T\sigma_{0}}{\sqrt{K}}+\frac{2\sqrt{2}\eta^{2}TL_{0}}{\sqrt{k(1-\beta)}}+\sum_{s=1}^{T}\sum_{t=s}^{T}2\eta\left(\sqrt{\frac{2}{k}}\eta L_{1}+\frac{(1-\beta)\sigma_{1}}{\sqrt{K}}\right)\beta^{t-s}\E\left[\|\na F(x_{s})\|\right]\nonumber \\
\leq & \frac{2\eta(\sigma_{0}+\sigma_{1}\|\na F(x_{1})\|)}{(1-\beta)\sqrt{K}}+\frac{2\eta\sqrt{1-\beta}T\sigma_{0}}{\sqrt{K}}+\frac{2\sqrt{2}\eta^{2}TL_{0}}{\sqrt{k(1-\beta)}}+\sum_{t=1}^{T}\eta\left(\frac{2\sqrt{2}\eta L_{1}}{\sqrt{k}(1-\beta)}+\frac{2\sigma_{1}}{\sqrt{K}}\right)\E\left[\|\na F(x_{t})\|\right].\label{eq:storm-exp-2}
\end{align}
Plugging (\ref{eq:storm-exp-2}) into (\ref{eq:storm-exp-1}) to obtain
\begin{align*}
 & \E\left[\Delta_{T+1}\right]+\sum_{t=1}^{T}\eta\E\left[\|\na F(x_{t})\|\right]\\
\leq & \Delta_{1}+\left(\frac{1}{2}+\frac{2\sqrt{2}}{\sqrt{k(1-\beta)}}\right)\eta^{2}TL_{0}+\frac{2\eta(\sigma_{0}+\sigma_{1}\|\na F(x_{1})\|)}{(1-\beta)\sqrt{K}}+\frac{2\eta\sqrt{1-\beta}T\sigma_{0}}{\sqrt{K}}\\
 & +\sum_{t=1}^{T}\eta\left(\frac{4\sqrt{\frac{2}{k}}+1-\beta}{2(1-\beta)}\eta L_{1}+\frac{2\sigma_{1}}{\sqrt{K}}\right)\E\left[\|\na F(x_{t})\|\right]\\
\leq & \Delta_{1}+\left(\frac{1}{2}+\frac{2\sqrt{2}}{\sqrt{k(1-\beta)}}\right)\eta^{2}TL_{0}+\frac{2\eta(\sigma_{0}+\sigma_{1}\|\na F(x_{1})\|)}{(1-\beta)\sqrt{K}}+\frac{2\eta\sqrt{1-\beta}T\sigma_{0}}{\sqrt{K}}+\sum_{t=1}^{T}\frac{\eta}{2}\E\left[\|\na F(x_{t})\|\right],
\end{align*}
where the last inequality is by $\eta\leq\frac{1-\beta}{2(4\sqrt{\frac{2}{k}}+1-\beta)L_{1}}\Rightarrow\frac{4\sqrt{\frac{2}{k}}+1-\beta}{2(1-\beta)}\eta L_{1}\leq\frac{1}{4}$
and $K\geq\left\lceil 64\sigma_{1}^{2}\right\rceil \Rightarrow\frac{2\sigma_{1}}{\sqrt{K}}\leq\frac{1}{4}$.
Then we know
\begin{align}
\E\left[\Delta_{T+1}\right]+\sum_{t=1}^{T}\frac{\eta}{2}\E\left[\|\na F(x_{t})\|\right] & \leq\Delta_{1}+\left(\frac{1}{2}+\frac{2\sqrt{2}}{\sqrt{k(1-\beta)}}\right)\eta^{2}TL_{0}+\frac{2\eta(\sigma_{0}+\sigma_{1}\|\na F(x_{1})\|)}{(1-\beta)\sqrt{K}}+\frac{2\eta\sqrt{1-\beta}T\sigma_{0}}{\sqrt{K}}\nonumber \\
\Rightarrow\sum_{t=1}^{T}\E\left[\|\na F(x_{t})\|\right]\leq & \frac{2\Delta_{1}}{\eta}+\left(1+\frac{4\sqrt{2}}{\sqrt{k(1-\beta)}}\right)\eta TL_{0}+\frac{4(\sigma_{0}+\sigma_{1}\|\na F(x_{1})\|)}{(1-\beta)\sqrt{K}}+\frac{4\sqrt{1-\beta}T\sigma_{0}}{\sqrt{K}}\nonumber \\
\overset{(a)}{\leq} & O\left(\Delta_{1}L_{1}+\sqrt{\Delta_{1}L_{0}T}+\frac{\sqrt{\Delta_{1}L_{0}T}}{\left(k(1-\beta)\right)^{1/4}}+\frac{\sqrt{1-\beta}T\sigma_{0}}{\sqrt{K}}\right)\nonumber \\
 & +O\left(\frac{(\sigma_{0}+\sigma_{1}\|\na F(x_{1})\|)/\sqrt{K}+\Delta_{1}L_{1}/\sqrt{k}}{1-\beta}\right)\nonumber \\
\overset{(b)}{\leq} & O\left(\Delta_{1}L_{1}+(\sigma_{0}+\sigma_{1}\|\na F(x_{1})\|)/\sqrt{K}+\sqrt{\Delta_{1}L_{0}T}\right)\nonumber \\
 & +O\left(\left(\frac{\sigma_{0}\Delta_{1}L_{0}T^{2}}{\sqrt{kK}}\right)^{\frac{1}{3}}+\left(\left(\frac{\sigma_{0}+\sigma_{1}\|\na F(x_{1})\|}{K^{3/2}}+\frac{\Delta_{1}L_{1}}{\sqrt{k}K}\right)\sigma_{0}^{2}T^{2}\right)^{\frac{1}{3}}\right).\label{eq:storm-exp-3}
\end{align}
where $(a)$ and $(b)$ are due to
\begin{align*}
\eta & =\min\left\{ \sqrt{\frac{\Delta_{1}\min\left\{ \sqrt{k(1-\beta)},1\right\} }{TL_{0}}},\frac{1-\beta}{2(4\sqrt{\frac{2}{k}}+1-\beta)L_{1}}\right\} ;\\
1-\beta & =\min\left\{ 1,\max\left\{ \left(\frac{\Delta_{1}L_{0}K}{\sigma_{0}^{2}\sqrt{k}T}\right)^{\frac{2}{3}},\left(\frac{\sigma_{0}+\sigma_{1}\|\na F(x_{1})\|+\Delta_{1}L_{1}\sqrt{K/k}}{\sigma_{0}T}\right)^{\frac{2}{3}}\right\} \right\} .
\end{align*}
(\ref{eq:storm-exp-3}) immediately implies
\begin{align*}
\min_{t\in\left[T\right]}\E\left[\|\na F(x_{t})\|\right]\leq & O\left(\frac{\Delta_{1}L_{1}+(\sigma_{0}+\sigma_{1}\|\na F(x_{1})\|)/\sqrt{K}}{T}+\sqrt{\frac{\Delta_{1}L_{0}}{T}}\right.\\
 & +\left.\sqrt[3]{\frac{\sigma_{0}\Delta_{1}L_{0}}{\sqrt{kK}T}+\frac{\sigma_{0}^{2}(\sigma_{0}+\sigma_{1}\|\na F(x_{1})\|)}{K^{3/2}T}+\frac{\sigma_{0}^{2}\Delta_{1}L_{1}}{\sqrt{k}KT}}\right).
\end{align*}
\end{proof}

\section{Missing Proofs in Section \ref{sec:analysis}\label{sec:app-analysis}}

In this section, we aim to prove all lemmas stated in Section \ref{sec:analysis}.
First recall our nations as follows
\begin{align*}
\Delta_{t\in\left[T\right]} & =F(x_{t})-F_{*};\\
\epsilon_{t\in\left\{ 0\right\} \cup\left[T\right]} & =\begin{cases}
\na f(x_{1},K_{1})-\na F(x_{1}) & t=0\\
m_{t}-\na F(x_{t}) & t\in\left[T\right]
\end{cases};\\
Z_{t\in\left[T\right]} & =\mathds{1}_{t\geq2}(\na f(x_{t},k_{t})-\na f(x_{t-1},k_{t})-\na F(x_{t})+\na F(x_{t-1}));\\
\xi_{t\in\left[T\right]} & =\na f(x_{t},K_{t})-\na F(x_{t}).
\end{align*}
Additionally, $\F_{t}$ is the natural filtration generated by $\left\{ K_{s}=\left\{ \Xi_{s}^{i}:i\in\left[K\right]\right\} ,\forall s\in\left[t\right]\right\} $.
Note that $x_{t}$ is $\F_{t-1}$ measurable, $Z_{t}$ and $\xi_{t}$
are both adapted to $\F_{t}$.

\subsection{Proof of Lemma \ref{lem:descent-lemma-any-time}}

\begin{proof}
If $t=0$, the inequality holds automatically. Now suppose $t\geq1$,
note that $\|x_{s+1}-x_{s}\|=\eta\leq\frac{1}{L_{1}}$ for any $s\in\left[T\right]$.
Hence, by Lemma \ref{lem:descent-lemma}, there is
\begin{align}
F(x_{s+1}) & \leq F(x_{s})+\langle\na F(x_{s}),x_{s+1}-x_{s}\rangle+\frac{L_{0}+L_{1}\|\na F(x_{s})\|}{2}\|x_{s+1}-x_{s}\|^{2}\nonumber \\
 & =F(x_{s})-\eta\langle\na F(x_{s}),\frac{m_{s}}{\|m_{s}\|}\rangle+\eta^{2}\frac{L_{0}+L_{1}\|\na F(x_{s})\|}{2}\nonumber \\
 & =F(x_{s})-\eta\|m_{s}\|+\eta\langle\epsilon_{s},\frac{m_{s}}{\|m_{s}\|}\rangle+\eta^{2}\frac{L_{0}+L_{1}\|\na F(x_{s})\|}{2}\nonumber \\
 & \overset{(a)}{\leq}F(x_{s})-\eta\|m_{s}\|+\eta\|\epsilon_{s}\|+\eta^{2}\frac{L_{0}+L_{1}\|\na F(x_{s})\|}{2}\nonumber \\
 & \overset{(b)}{\leq}F(x_{s})-\eta\|\na F(x_{s})\|+2\eta\|\epsilon_{s}\|+\eta^{2}\frac{L_{0}+L_{1}\|\na F(x_{s})\|}{2}\nonumber \\
\Rightarrow\Delta_{s+1}+\eta\|\na F(x_{s})\|\leq & \Delta_{s}+2\eta\|\epsilon_{s}\|+\eta^{2}\frac{L_{0}+L_{1}\|\na F(x_{s})\|}{2},\label{eq:descent-lemma-any-time-1}
\end{align}
where $(a)$ is by Cauchy-Schwarz inequality and $(b)$ is by $\|m_{s}\|=\|\na F(x_{s})+\epsilon_{s}\|\geq\|\na F(x_{s})\|-\|\epsilon_{s}\|$.
Summing up (\ref{eq:descent-lemma-any-time-1}) from $s=1$ to $t$
to get the desired result.
\end{proof}

\subsection{Proof of Lemma \ref{lem:decomposition}}

\begin{proof}
We use the definitions of $m_{t}$, $\epsilon_{t}$, $Z_{t}$ and
$\xi_{t}$ here. For $t\geq2$
\begin{align*}
\epsilon_{t} & =m_{t}-\nabla F(x_{t})\\
 & =\beta m_{t-1}+(1-\beta)\na f(x_{t},K_{t})+\beta(\na f(x_{t},k_{t})-\na f(x_{t-1},k_{t}))-\na F(x_{t})\\
 & =\beta\epsilon_{t-1}+\beta Z_{t}+(1-\beta)\xi_{t}.
\end{align*}
Note that the above equation also holds when $t=1$. By expanding
the recursion, we can obtain
\[
\epsilon_{t}=\beta^{t}\epsilon_{0}+\beta\sum_{s=1}^{t}\beta^{t-s}Z_{s}+(1-\beta)\sum_{s=1}^{t}\beta^{t-s}\xi_{s}.
\]
\end{proof}

\subsection{Proof of Lemma \ref{lem:hp-bound-Z}}

\begin{proof}
Note that $\beta^{t-s}Z_{s}\mathds{1}\left[\|\na F(x_{s})\|\leq G\right]\in\F_{s},\forall s\in\left[t\right]$.
Additionally, we observe that by Assumption 2
\[
\E\left[\beta^{t-s}Z_{s}\mathds{1}\left[\|\na F(x_{s})\|\leq G\right]\mid\F_{s-1}\right]=\beta^{t-s}\mathds{1}\left[\|\na F(x_{s})\|\leq G\right]\E\left[Z_{s}\mid\F_{s-1}\right]=0.
\]
Besides, by Assumption 3A, we know for $s\geq2$
\begin{align*}
 & \|\beta^{t-s}Z_{s}\mathds{1}\left[\|\na F(x_{s})\|\leq G\right]\|\\
\leq & \beta^{t-s}\mathds{1}\left[\|\na F(x_{s})\|\leq G\right]\|Z_{s}\|\\
\leq & \beta^{t-s}\mathds{1}\left[\|\na F(x_{s})\|\leq G\right]\left(\|\na f(x_{s},k_{s})-\na f(x_{s-1},k_{s})\|+\|\na F(x_{s})-\na F(x_{s-1})\|\right)\\
\leq & \beta^{t-s}\mathds{1}\left[\|\na F(x_{s})\|\leq G\right]\left(\frac{1}{k}\sum_{i=1}^{k}\|\na f(x_{s},\Xi_{s}^{i})-\na f(x_{s-1},\Xi_{s}^{i})\|+\|\na F(x_{s})-\na F(x_{s-1})\|\right)\\
\leq & 2\beta^{t-s}\mathds{1}\left[\|\na F(x_{s})\|\leq G\right]\left(L_{0}+L_{1}\|\na F(x_{s})\|\right)\|x_{s}-x_{s-1}\|\\
\leq & 2\beta^{t-s}(L_{0}+L_{1}G)\eta.
\end{align*}
and $||\beta^{t-1}Z_{1}\mathds{1}\left[\|\na F(x_{1})\|\leq G\right]\|=0\leq2\beta^{t-1}(L_{0}+L_{1}G)\eta$.

Thus, $\beta^{t-s}Z_{s}\mathds{1}\left[\|\na F(x_{s})\|\leq G\right],\forall s\in\left[t\right]$
is a bounded martingale difference sequence. Now we apply Lemma \ref{lem:concentration}
to obtain with probability at least $1-\delta$, for any $\tau\in\left[t\right]$,
there is
\[
\left\Vert \sum_{s=1}^{\tau}\beta^{t-s}Z_{s}\mathds{1}\left[\|\na F(x_{s})\|\leq G\right]\right\Vert \leq4\sqrt{\log\frac{2}{\delta}\sum_{s=1}^{t}\left(2\beta^{t-s}(L_{0}+L_{1}G)\eta\right)^{2}}\leq8\eta(L_{0}+L_{1}G)\sqrt{\frac{\log\frac{2}{\delta}}{1-\beta}},
\]
which implies with probability at least $1-\delta$, we have
\[
\left\Vert \sum_{s=1}^{t}\beta^{t-s}Z_{s}\mathds{1}\left[\|\na F(x_{s})\|\leq G\right]\right\Vert \leq8\eta(L_{0}+L_{1}G)\sqrt{\frac{\log\frac{2}{\delta}}{1-\beta}}.
\]
By replacing $\delta$ with $\frac{\delta}{2T}$, we finish the proof.
\end{proof}

\subsection{Proof of Lemma \ref{lem:hp-bound-xi}}

\begin{proof}
Note that $\beta^{t-s}\xi_{s}\mathds{1}\left[\|\na F(x_{s})\|\leq G\right]\in\F_{s},\forall s\in\left[t\right]$.
Additionally, we observe that by Assumption 2
\[
\E\left[\beta^{t-s}\xi_{s}\mathds{1}\left[\|\na F(x_{s})\|\leq G\right]\mid\F_{s-1}\right]=\beta^{t-s}\mathds{1}\left[\|\na F(x_{s})\|\leq G\right]\E\left[\xi_{s}\mid\F_{s-1}\right]=0.
\]
Besides, by Assumption 4A, we know for $s\in\left[t\right]$
\begin{align*}
 & \|\beta^{t-s}\xi_{s}\mathds{1}\left[\|\na F(x_{s})\|\leq G\right]\|\\
\leq & \beta^{t-s}\mathds{1}\left[\|\na F(x_{s})\|\leq G\right]\|\xi_{s}\|\\
\leq & \beta^{t-s}\mathds{1}\left[\|\na F(x_{s})\|\leq G\right]\frac{1}{K}\sum_{i=1}^{K}\|\na f(x_{s},\Xi_{s}^{i})-\na F(x_{s})\|\\
\leq & \beta^{t-s}\mathds{1}\left[\|\na F(x_{s})\|\leq G\right]\left(\sigma_{0}+\sigma_{1}\|\na F(x_{s})\|\right)\\
\leq & \beta^{t-s}(\sigma_{0}+\sigma_{1}G).
\end{align*}

Thus, $\beta^{t-s}\xi_{s}\mathds{1}\left[\|\na F(x_{s})\|\leq G\right],\forall s\in\left[t\right]$
is a bounded martingale difference sequence. Now we apply Lemma \ref{lem:concentration}
to obtain with probability at least $1-\delta$, for any $\tau\in\left[t\right]$,
there is
\[
\left\Vert \sum_{s=1}^{\tau}\beta^{t-s}\xi_{s}\mathds{1}\left[\|\na F(x_{s})\|\leq G\right]\right\Vert \leq4\sqrt{\log\frac{2}{\delta}\sum_{s=1}^{t}\left(\beta^{t-s}(\sigma_{0}+\sigma_{1}G)\right)^{2}}\leq4(\sigma_{0}+\sigma_{1}G)\sqrt{\frac{\log\frac{2}{\delta}}{1-\beta}},
\]
which implies with probability at least $1-\delta$, we have
\[
\left\Vert \sum_{s=1}^{t}\beta^{t-s}\xi_{s}\mathds{1}\left[\|\na F(x_{s})\|\leq G\right]\right\Vert \leq4(\sigma_{0}+\sigma_{1}G)\sqrt{\frac{\log\frac{2}{\delta}}{1-\beta}}.
\]
By replacing $\delta$ with $\frac{\delta}{2T}$, we finish the proof.
\end{proof}

\subsection{Proof of Lemma \ref{lem:exp-bound-Z}}

\begin{proof}
If $t=1$, we have 
\[
\E\left[\left\Vert \sum_{s=1}^{t}\beta^{t-s}Z_{s}\right\Vert \right]=\E\left[\|Z_{1}\|\right]=0\leq\frac{\sqrt{2}\eta L_{0}}{\sqrt{k(1-\beta)}}+\sqrt{\frac{2}{k}}\eta L_{1}\E\left[\|\na F(x_{1})\|\right].
\]
Now suppose $t\geq2$, we will prove the following result by induction:
for any $r\in\left\{ 0\right\} \cup\left[t\right]$,
\begin{equation}
\E\left[\left\Vert \sum_{s=1}^{t}\beta^{t-s}Z_{s}\right\Vert \mid\F_{t-r}\right]\leq\sqrt{\left\Vert \sum_{s=1}^{t-r}\beta^{t-s}Z_{s}\right\Vert ^{2}+\sum_{s=1}^{r}\frac{2\beta^{2s-2}\eta^{2}L_{0}^{2}}{k}}+\sum_{s=t+1-r}^{t}\sqrt{\frac{2}{k}}\eta L_{1}\beta^{t-s}\E\left[\|\na F(x_{s})\|\mid\F_{t-r}\right].\label{eq:exp-bound-Z-hypothesis}
\end{equation}

First, for the case $r=0$, we have
\[
\E\left[\left\Vert \sum_{s=1}^{t}\beta^{t-s}Z_{s}\right\Vert \mid\F_{t}\right]=\left\Vert \sum_{s=1}^{t}\beta^{t-s}Z_{s}\right\Vert =\sqrt{\left\Vert \sum_{s=1}^{t}\beta^{t-s}Z_{s}\right\Vert ^{2}},
\]
which means (\ref{eq:exp-bound-Z-hypothesis}) holds. Suppose (\ref{eq:exp-bound-Z-hypothesis})
holds for $r=r_{0}\in\left[t-1\right]$. For $r=r_{0}+1$, we have
\begin{align*}
 & \E\left[\left\Vert \sum_{s=1}^{t}\beta^{t-s}Z_{s}\right\Vert \mid\F_{t-r_{0}-1}\right]\\
= & \E\left[\E\left[\left\Vert \sum_{s=1}^{t}\beta^{t-s}Z_{s}\right\Vert \mid\F_{t-r_{0}}\right]\mid\F_{t-r_{0}-1}\right]\\
\leq & \E\left[\sqrt{\left\Vert \sum_{s=1}^{t-r_{0}}\beta^{t-s}Z_{s}\right\Vert ^{2}+\sum_{s=1}^{r_{0}}\frac{2\beta^{2s-2}\eta^{2}L_{0}^{2}}{k}}+\sum_{s=t+1-r_{0}}^{t}\sqrt{\frac{2}{k}}\eta L_{1}\beta^{t-s}\E\left[\|\na F(x_{s})\|\mid\F_{t-r_{0}}\right]\mid\F_{t-r_{0}-1}\right]\\
= & \E\left[\sqrt{\left\Vert \sum_{s=1}^{t-r_{0}}\beta^{t-s}Z_{s}\right\Vert ^{2}+\sum_{s=1}^{r_{0}}\frac{2\beta^{2s-2}\eta^{2}L_{0}^{2}}{k}}\mid\F_{t-r_{0}-1}\right]+\sum_{s=t+1-r_{0}}^{t}\sqrt{\frac{2}{k}}\eta L_{1}\beta^{t-s}\E\left[\|\na F(x_{s})\|\mid\F_{t-r_{0}-1}\right]\\
\overset{(c)}{\leq} & \sqrt{\E\left[\left\Vert \sum_{s=1}^{t-r_{0}}\beta^{t-s}Z_{s}\right\Vert ^{2}+\sum_{s=1}^{r_{0}}\frac{2\beta^{2s-2}\eta^{2}L_{0}^{2}}{k}\mid\F_{t-r_{0}-1}\right]}+\sum_{s=t+1-r_{0}}^{t}\sqrt{\frac{2}{k}}\eta L_{1}\beta^{t-s}\E\left[\|\na F(x_{s})\|\mid\F_{t-r_{0}-1}\right]\\
= & \sqrt{\beta^{2r_{0}}\E\left[\|Z_{t-r_{0}}\|^{2}\mid\F_{t-r_{0}-1}\right]+2\left\langle \E\left[\beta^{r_{0}}Z_{t-r_{0}}\mid\F_{t-r_{0}-1}\right],\sum_{s=1}^{t-r_{0}-1}\beta^{t-s}Z_{s}\right\rangle +\left\Vert \sum_{s=1}^{t-r_{0}-1}\beta^{t-s}Z_{s}\right\Vert ^{2}+\sum_{s=1}^{r_{0}}\frac{2\beta^{2s-2}\eta^{2}L_{0}^{2}}{k}}\\
 & +\sum_{s=t+1-r_{0}}^{t}\sqrt{\frac{2}{k}}\beta^{t-s}\eta L_{1}\E\left[\|\na F(x_{s})\|\mid\F_{t-r_{0}-1}\right]\\
\overset{(d)}{\leq} & \sqrt{\beta^{2r_{0}}\frac{(L_{0}+L_{1}\|\na F(x_{t-r_{0}})\|)^{2}\eta^{2}}{k}+\left\Vert \sum_{s=1}^{t-r_{0}-1}\beta^{t-s}Z_{s}\right\Vert ^{2}+\sum_{s=1}^{r_{0}}\frac{2\beta^{2s-2}\eta^{2}L_{0}^{2}}{k}}\\
 & +\sum_{s=t+1-r_{0}}^{t}\sqrt{\frac{2}{k}}\eta L_{1}\beta^{t-s}\E\left[\|\na F(x_{s})\|\mid\F_{t-r_{0}-1}\right]\\
\leq & \sqrt{\left\Vert \sum_{s=1}^{t-r_{0}-1}\beta^{t-s}Z_{s}\right\Vert ^{2}+\sum_{s=1}^{r_{0}+1}\frac{2\beta^{2s-2}\eta^{2}L_{0}^{2}}{k}}+\sum_{s=t-r_{0}}^{t}\sqrt{\frac{2}{k}}\eta L_{1}\beta^{t-s}\E\left[\|\na F(x_{s})\|\mid\F_{t-r_{0}-1}\right],
\end{align*}
where $(c)$ is by Holder inequality; for $(d)$, we first note that
$\E\left[\beta^{r_{0}}Z_{t-r_{0}}\mid\F_{t-r_{0}-1}\right]=0$ is
true. If $t-r_{0}\geq2$
\begin{align*}
 & \E\left[\|Z_{t-r_{0}}\|^{2}\mid\F_{t-r_{0}-1}\right]\\
= & \E\left[\|\na f(x_{t-r_{0}},k_{t-r_{0}})-\na f(x_{t-r_{0}-1},k_{t-r_{0}})-\na F(x_{t-r_{0}})+\na F(x_{t-r_{0}-1}))\|^{2}\mid\F_{t-r_{0}-1}\right]\\
\leq & \E\left[\|\na f(x_{t-r_{0}},k_{t-r_{0}})-\na f(x_{t-r_{0}-1},k_{t-r_{0}})\|^{2}\mid\F_{t-r_{0}-1}\right]\\
= & \frac{1}{k^{2}}\sum_{i=1}^{k}\E\left[\|\na f(x_{t-r_{0}},\Xi_{t-r_{0}}^{i})-\na f(x_{t-r_{0}-1},\Xi_{t-r_{0}}^{i})\|^{2}\mid\F_{t-r_{0}-1}\right]\\
\leq & \frac{(L_{0}+L_{1}\|\na F(x_{t-r_{0}})\|)^{2}\|x_{t-r_{0}}-x_{t-r_{0}-1}\|^{2}}{k}\\
= & \frac{(L_{0}+L_{1}\|\na F(x_{t-r_{0}})\|)^{2}\eta^{2}}{k},
\end{align*}
where the last inequality holds due to Assumption 3B; for $t-r_{0}=1$,
we know
\[
\E\left[\|Z_{t-r_{0}}\|^{2}\mid\F_{t-r_{0}-1}\right]=\E\left[\|Z_{1}\|^{2}\right]=0\leq\frac{(L_{0}+L_{1}\|\na F(x_{1})\|)^{2}\eta^{2}}{k}.
\]
Hence, by induction, (\ref{eq:exp-bound-Z-hypothesis}) holds for
$r\in\left\{ 0\right\} \cup\left[t\right]$. In particular, taking
$r=t$, we obtain the following bound
\begin{align*}
\E\left[\left\Vert \sum_{s=1}^{t}\beta^{t-s}Z_{s}\right\Vert \right] & \leq\sqrt{\sum_{s=1}^{t}\frac{2\beta^{2s-2}\eta^{2}L_{0}^{2}}{k}}+\sum_{s=1}^{t}\sqrt{\frac{2}{k}}\eta L_{1}\beta^{t-s}\E\left[\|\na F(x_{s})\|\right]\\
 & \leq\frac{\sqrt{2}\eta L_{0}}{\sqrt{k(1-\beta^{2})}}+\sum_{s=1}^{t}\sqrt{\frac{2}{k}}\eta L_{1}\beta^{t-s}\E\left[\|\na F(x_{s})\|\right]\\
 & \leq\frac{\sqrt{2}\eta L_{0}}{\sqrt{k(1-\beta)}}+\sum_{s=1}^{t}\sqrt{\frac{2}{k}}\eta L_{1}\beta^{t-s}\E\left[\|\na F(x_{s})\|\right].
\end{align*}
\end{proof}

\subsection{Proof of Lemma \ref{lem:exp-bound-xi}}

\begin{proof}
We will prove the following result by induction: for any $r\in\left\{ 0\right\} \cup\left[t\right]$,
\begin{equation}
\E\left[\left\Vert \sum_{s=1}^{t}\beta^{t-s}\xi_{s}\right\Vert \mid\F_{t-r}\right]\leq\sqrt{\left\Vert \sum_{s=1}^{t-r}\beta^{t-s}\xi_{s}\right\Vert ^{2}+\sum_{s=1}^{r}\frac{\beta^{2s-2}\sigma_{0}^{2}}{K}}+\sum_{s=t+1-r}^{t}\frac{\sigma_{1}}{\sqrt{K}}\beta^{t-s}\E\left[\|\na F(x_{s})\|\mid\F_{t-r}\right].\label{eq:exp-bound-xi-hypothesis}
\end{equation}

First, for the case $r=0$, we have
\[
\E\left[\left\Vert \sum_{s=1}^{t}\beta^{t-s}\xi_{s}\right\Vert \mid\F_{t}\right]=\left\Vert \sum_{s=1}^{t}\beta^{t-s}\xi_{s}\right\Vert =\sqrt{\left\Vert \sum_{s=1}^{t}\beta^{t-s}\xi_{s}\right\Vert ^{2}},
\]
which means (\ref{eq:exp-bound-xi-hypothesis}) holds. Suppose (\ref{eq:exp-bound-xi-hypothesis})
holds for $r=r_{0}\in\left[t-1\right]$. For $r=r_{0}+1$, we have
\begin{align*}
 & \E\left[\left\Vert \sum_{s=1}^{t}\beta^{t-s}\xi_{s}\right\Vert \mid\F_{t-r_{0}-1}\right]\\
= & \E\left[\E\left[\left\Vert \sum_{s=1}^{t}\beta^{t-s}\xi_{s}\right\Vert \mid\F_{t-r_{0}}\right]\mid\F_{t-r_{0}-1}\right]\\
\leq & \E\left[\sqrt{\left\Vert \sum_{s=1}^{t-r_{0}}\beta^{t-s}\xi_{s}\right\Vert ^{2}+\sum_{s=1}^{r_{0}}\frac{\beta^{2s-2}\sigma_{0}^{2}}{K}}+\sum_{s=t+1-r_{0}}^{t}\frac{\sigma_{1}}{\sqrt{K}}\beta^{t-s}\E\left[\|\na F(x_{s})\|\mid\F_{t-r_{0}}\right]\mid\F_{t-r_{0}-1}\right]\\
= & \E\left[\sqrt{\left\Vert \sum_{s=1}^{t-r_{0}}\beta^{t-s}\xi_{s}\right\Vert ^{2}+\sum_{s=1}^{r_{0}}\frac{\beta^{2s-2}\sigma_{0}^{2}}{K}}\mid\F_{t-r_{0}-1}\right]+\sum_{s=t+1-r_{0}}^{t}\frac{\sigma_{1}}{\sqrt{K}}\beta^{t-s}\E\left[\|\na F(x_{s})\|\mid\F_{t-r_{0}-1}\right]\\
\overset{(c)}{\leq} & \sqrt{\E\left[\left\Vert \sum_{s=1}^{t-r_{0}}\beta^{t-s}\xi_{s}\right\Vert ^{2}+\sum_{s=1}^{r_{0}}\frac{\beta^{2s-2}\sigma_{0}^{2}}{K}\mid\F_{t-r_{0}-1}\right]}+\sum_{s=t+1-r_{0}}^{t}\frac{\sigma_{1}}{\sqrt{K}}\beta^{t-s}\E\left[\|\na F(x_{s})\|\mid\F_{t-r_{0}-1}\right]\\
= & \sqrt{\beta^{2r_{0}}\E\left[\|\xi_{t-r_{0}}\|^{2}\mid\F_{t-r_{0}-1}\right]+2\left\langle \E\left[\beta^{r_{0}}\xi_{t-r_{0}}\mid\F_{t-r_{0}-1}\right],\sum_{s=1}^{t-r_{0}-1}\beta^{t-s}\xi_{s}\right\rangle +\left\Vert \sum_{s=1}^{t-r_{0}-1}\beta^{t-s}\xi_{s}\right\Vert ^{2}+\sum_{s=1}^{r_{0}}\frac{\beta^{2s-2}\sigma_{0}^{2}}{K}}\\
 & +\sum_{s=t+1-r_{0}}^{t}\frac{\sigma_{1}}{\sqrt{K}}\beta^{t-s}\eta L_{1}\E\left[\|\na F(x_{s})\|\mid\F_{t-r_{0}-1}\right]\\
\overset{(d)}{\leq} & \sqrt{\beta^{2r_{0}}\frac{\sigma_{0}^{2}+\sigma_{1}^{2}\|\na F(x_{t-r_{0}})\|^{2}}{K}+\left\Vert \sum_{s=1}^{t-r_{0}-1}\beta^{t-s}\xi_{s}\right\Vert ^{2}+\sum_{s=1}^{r_{0}}\frac{\beta^{2s-2}\sigma_{0}^{2}}{K}}\\
 & +\sum_{s=t+1-r_{0}}^{t}\frac{\sigma_{1}}{\sqrt{K}}\beta^{t-s}\eta L_{1}\E\left[\|\na F(x_{s})\|\mid\F_{t-r_{0}-1}\right]\\
\leq & \sqrt{\left\Vert \sum_{s=1}^{t-r_{0}-1}\beta^{t-s}\xi_{s}\right\Vert ^{2}+\sum_{s=1}^{r_{0}+1}\frac{\beta^{2s-2}\sigma_{0}^{2}}{K}}+\sum_{s=t-r_{0}}^{t}\frac{\sigma_{1}}{\sqrt{K}}\beta^{t-s}\E\left[\|\na F(x_{s})\|\mid\F_{t-r_{0}-1}\right],
\end{align*}
where $(c)$ is by Holder inequality, $(d)$ is due to we first note
that $\E\left[\beta^{r_{0}}\xi_{t-r_{0}}\mid\F_{t-r_{0}-1}\right]=0$
and
\begin{align*}
 & \E\left[\|\xi_{t-r_{0}}\|^{2}\mid\F_{t-r_{0}-1}\right]\\
= & \E\left[\|\na f(x_{t-r_{0}},K_{t-r_{0}})-\na F(x_{t-r_{0}})\|^{2}\mid\F_{t-r_{0}-1}\right]\\
= & \frac{1}{K^{2}}\sum_{i=1}^{K}\E\left[\|\na f(x_{t-r_{0}},\Xi_{t-r_{0}}^{i})-\na F(x_{t-r_{0}})\|^{2}\mid\F_{t-r_{0}-1}\right]\\
\leq & \frac{\sigma_{0}^{2}+\sigma_{1}^{2}\|\na F(x_{t-r_{0}})\|^{2}}{K},
\end{align*}
where the last inequality holds due to Assumption 4B.

Hence, by induction, (\ref{eq:exp-bound-xi-hypothesis}) holds for
$r\in\left\{ 0\right\} \cup\left[t\right]$. In particular, taking
$r=t$, we obtain the following bound
\begin{align*}
\E\left[\left\Vert \sum_{s=1}^{t}\beta^{t-s}\xi_{s}\right\Vert \right] & \leq\sqrt{\sum_{s=1}^{t}\frac{\beta^{2s-2}\sigma_{0}^{2}}{K}}+\sum_{s=1}^{t}\frac{\sigma_{1}}{\sqrt{K}}\beta^{t-s}\E\left[\|\na F(x_{s})\|\right]\\
 & \leq\frac{\sigma_{0}}{\sqrt{K(1-\beta^{2})}}+\sum_{s=1}^{t}\frac{\sigma_{1}}{\sqrt{K}}\beta^{t-s}\E\left[\|\na F(x_{s})\|\right]\\
 & \leq\frac{\sigma_{0}}{\sqrt{K(1-\beta)}}+\sum_{s=1}^{t}\frac{\sigma_{1}}{\sqrt{K}}\beta^{t-s}\E\left[\|\na F(x_{s})\|\right].
\end{align*}
\end{proof}

\subsection{Proof of Lemma \ref{lem:exp-bound-err}}

\begin{proof}
By Lemma \ref{lem:decomposition}, we know
\begin{align}
\|\epsilon_{t}\| & \leq\beta^{t}\|\epsilon_{0}\|+\beta\left\Vert \sum_{s=1}^{t}\beta^{t-s}Z_{s}\right\Vert +(1-\beta)\left\Vert \sum_{s=1}^{t}\beta^{t-s}\xi_{s}\right\Vert \nonumber \\
\Rightarrow\E\left[\|\epsilon_{t}\|\right] & \leq\beta^{t}\E\left[\|\epsilon_{0}\|\right]+\beta\E\left[\left\Vert \sum_{s=1}^{t}\beta^{t-s}Z_{s}\right\Vert \right]+(1-\beta)\E\left[\left\Vert \sum_{s=1}^{t}\beta^{t-s}\xi_{s}\right\Vert \right]\nonumber \\
 & \leq\beta^{t}\E\left[\|\epsilon_{0}\|\right]+\E\left[\left\Vert \sum_{s=1}^{t}\beta^{t-s}Z_{s}\right\Vert \right]+(1-\beta)\E\left[\left\Vert \sum_{s=1}^{t}\beta^{t-s}\xi_{s}\right\Vert \right].\label{eq:exp-bound-err-1}
\end{align}

Note that
\begin{align}
\E\left[\|\epsilon_{0}\|\right] & =\E\left[\|\nabla f(x_{1},K_{1})-\na F(x_{1})\|\right]\overset{(a)}{\leq}\sqrt{\E\left[\|\nabla f(x_{1},K_{1})-\na F(x_{1})\|^{2}\right]}\nonumber \\
 & \overset{(b)}{\leq}\sqrt{\frac{\sigma_{0}^{2}+\sigma_{1}^{2}\|\na F(x_{1})\|^{2}}{K}}\leq\frac{\sigma_{0}+\sigma_{1}\|\na F(x_{1})\|}{\sqrt{K}},\label{eq:exp-bound-err-2}
\end{align}
where $(a)$ is by Holder inequality, $(b)$ is by Assumption 4B and
$\Xi_{1}^{i},i\in\left[K\right]$ are independent.

By Lemmas \ref{lem:exp-bound-Z} and \ref{lem:exp-bound-xi}, there
are
\begin{align}
\E\left[\left\Vert \sum_{s=1}^{t}\beta^{t-s}Z_{s}\right\Vert \right] & \leq\frac{\sqrt{2}\eta L_{0}}{\sqrt{k(1-\beta)}}+\sum_{s=1}^{t}\sqrt{\frac{2}{k}}\eta L_{1}\beta^{t-s}\E\left[\|\na F(x_{s})\|\right],\label{eq:exp-bound-err-3}\\
\E\left[\left\Vert \sum_{s=1}^{t}\beta^{t-s}\xi_{s}\right\Vert \right] & \leq\frac{\sigma_{0}}{\sqrt{K(1-\beta)}}+\sum_{s=1}^{t}\frac{\sigma_{1}}{\sqrt{K}}\beta^{t-s}\E\left[\|\na F(x_{s})\|\right].\label{eq:exp-bound-err-4}
\end{align}

Finally, we plug (\ref{eq:exp-bound-err-2}), (\ref{eq:exp-bound-err-3})
and (\ref{eq:exp-bound-err-4}) into (\ref{eq:exp-bound-err-1}) to
obtain
\[
\E\left[\|\epsilon_{t}\|\right]\leq\beta^{t}\frac{\sigma_{0}+\sigma_{1}\|\na F(x_{1})\|}{\sqrt{K}}+\frac{\sqrt{1-\beta}\sigma_{0}}{\sqrt{K}}+\frac{\sqrt{2}\eta L_{0}}{\sqrt{k(1-\beta)}}+\sum_{s=1}^{t}\left(\sqrt{\frac{2}{k}}\eta L_{1}+\frac{(1-\beta)\sigma_{1}}{\sqrt{K}}\right)\beta^{t-s}\E\left[\|\na F(x_{s})\|\right].
\]
\end{proof}

\end{document}